%% file: main.tex
\documentclass{article}

\PassOptionsToPackage{hyphens}{url}
\PassOptionsToPackage{numbers,compress}{natbib}
\usepackage[sglblindworkshop]{neurips_2026}
\workshoptitle{AI at Scale for Clinical Impact (ASCI): Cancer Pathology Foundation Models}

\usepackage[utf8]{inputenc}
\usepackage[T1]{fontenc}
\usepackage{courier}
\usepackage{hyperref}

\renewcommand{\floatpagefraction}{0.75}
\hypersetup{hidelinks} 
\usepackage{url}
\usepackage{booktabs}
\usepackage{amsmath}
\usepackage{amssymb}
\usepackage{microtype}
\usepackage{xcolor}
\usepackage{graphicx}
\usepackage{flafter}
\usepackage{float}

\definecolor{worse}{HTML}{B91C1C}
\newcommand{\gain}[1]{\textbf{#1}}
\newcommand{\loss}[1]{\textcolor{worse}{#1}}

\title{Contrastive fine-tuning against stain and scanner bias improves robustness and downstream performance across pathology foundation models}

\author{%
  \textbf{Ryan Kim}$^{1,2}$ \quad \textbf{Hassan Muhammad}$^{1,3}$ \quad \textbf{Anish Dulal}$^{1,4}$ \\
  \textbf{Paul S. Scotti}$^{1,3}$ \quad \textbf{Tanishq Mathew Abraham}$^{1,3}$ \\[3pt]
  $^1$MedARC \quad $^2$University of Illinois at Urbana-Champaign \\
  $^3$Sophont \quad $^4$University of Oregon \\
  \texttt{tanishq@sophont.med}
}

\begin{document}

\maketitle

\begin{abstract}
Histopathology foundation models achieve strong performance across diverse downstream tasks, yet their representations have been shown to encode scanner- and site-specific biases, creating generalization gaps and potentially limiting real-world utility. We introduce \textsc{SPECTRA} (Scanner and Pathology Embedding Contrastive Tuning for Robust Alignment), a model-agnostic contrastive learning post-training framework which explicitly encourages scanner- and stain-robust representations, to optimize performance and generalization on downstream applications. Using parameter-efficient LoRA adapters trained on matched multi-scanner and multi-stain whole-slide images sourced from the PLISM dataset, our method aligns spatially corresponding regions across acquisition conditions (scanner, stain), using other locations within each candidate condition as negatives. \textsc{SPECTRA} improved PathoROB robustness index, gene-expression prediction on HEST, slide-level classification AUC on CPTAC, and held-out PLISM retrieval for every model. Overall tile-level classification on THUNDER also improved, with exceptions for segmentation and calibration tasks.  We open-source our training and evaluation codebases and release adapters and model weights for selected backbones.
\end{abstract}

\section{Introduction}

A digital pathology-based cancer diagnosis, biomarker prediction, or prognostic assessment should not be biased to the laboratory which prepares the reference tissue specimen or the scanner which digitizes it. Pathology foundation models promise to support diverse clinical tasks, yet strong benchmark performance alone does not establish reliability across real-world applications and their related deployment environments \citep{chen2024uni, huang2025flex}. Most leading models are pretrained on proprietary samples drawn from a single institution or a small number of health systems. Although these collections can be large, their limited institutional diversity may expose models to a narrow range of patient populations, laboratory protocols, stains, and scanner hardware. Even The Cancer Genome Atlas (TCGA), one of the largest public digital-pathology repositories, does not eliminate this risk: despite aggregating specimens across more than 140 geographically distributed institutions within a centralized digitization program, each tissue-source-site code denotes a specific institution--cancer-type pair, creating a perfect association between that code and the corresponding disease cohort, making models susceptible to recovering acquisition site from image features \citep{gdc_tss_codes,dehkharghanian2023biased,howard2021site}. Consequently, learned representations can entangle disease morphology with institution-specific acquisition signatures, allowing downstream adapters to exploit spurious correlations that fail when transferred to cohorts external to training data \citep{huang2025flex}.

Evidence of this vulnerability is increasingly widespread. PathoROB \citep{koemen2026pathorob}, which evaluates 20 foundation models in 34 medical centers, demonstrated persistent sensitivity to quantifiable non-biological variation. Conventional discrimination metrics can also conceal clinically relevant instability. For example, a controlled study of identical slides digitized on different devices found scanner-dependent calibration shifts even when AUC remained largely stable, with robustness not reliably predicted by model size or pretraining scale \citep{thiringer2026scanner}. In real-world practice, these shifts may move patients across diagnostic, prognostic, or biomarker-triage thresholds despite not having relevant tissue biology driving those associations. These findings motivate representation-learning methods which explicitly preserve disease-relevant informatio, while suppressing technical and non-biological variation, making representational stability across clinical environments a primary objective.

Motivated by this domain-wide problem, we introduce \textsc{SPECTRA}, a low-resource, model-agnostic fine-tuning framework that improves the robustness of pathology foundation-model representations to scanner and staining variation using parameter-efficient LoRA adapters and a label-free contrastive objective over spatially matched multi-scanner, multi-stain tiles. Training and evaluation code is available on \href{https://github.com/MedARC-AI/SPECTRA}{\textcolor{blue}{GitHub}}; model weights of fine-tuned models are available on \href{https://huggingface.co/collections/medarc/spectra-adapters-models}{\textcolor{blue}{Hugging Face}} for use in downstream pipelines.

\section{Related work}
\label{sec:related_work}

\paragraph{Robustness interventions in computational pathology.}
Existing approaches to addressing model robustness operate at several stages of the computational pathology pipeline. Image preprocessing and augmentation methods reduce color variation through stain normalization \citep{reinhard2001,macenko2009,vahadane2016structure} or expose models to broader appearance distributions through stain and color augmentation \citep{tellez2019quantifying,shen2022randstainna}. Although effective against some staining shifts, these methods do not explicitly address scanner-dependent differences in optical, resolution, compression, or image-processing software, and may suppress diagnostically relevant color information. Training-time approaches use domain-adversarial objectives \citep{ganin2016dann} or information-bottleneck formulations \citep{huang2025flex} to discourage prediction of acquisition domains, but generally require site labels and are optimized for a particular downstream task. Post-hoc methods instead harmonize frozen embeddings through feature alignment or batch correction \citep{nguyen2026fmmap,koemen2026pathorob}; these methods can improve robustness without modifying the encoder but may depend on target-domain statistics and may require adaptation to new deployment settings. Knowledge distillation offers another route to transfer invariant representations \citep{filiot2025distill}, although it requires a teacher and an additional distillation stage and introduces an additional computational cost. These approaches motivate reproducible, task-agnostic methods for adapting existing foundation-model encoders to scanner and staining variation.

\paragraph{Contrastive adaptation using matched scanners.}
To our knowledge, Carloni et al.\ \citep{carloni2025scangen} were the first to introduce contrastive fine-tuning for scanner generalization in pathology-foundation-model pipelines. Their ScanGen objective uses matched specimens digitized on different scanners as positive pairs and different specimens acquired on the same scanner as negative pairs. ScanGen trains a projection network placed between a frozen foundation-model encoder and a multiple-instance learning model, jointly optimizing scanner invariance and supervised EGFR-mutation prediction. The study provided an important proof of concept across five foundation models and a proprietary cohort of 323 lung-cancer specimens digitized on six scanners. However, its evaluation was confined to a single cancer type and biomarker, and robustness was assessed primarily through scanner-dependent variation in EGFR predictions. It did not evaluate whether the adapted representations retain general utility across standardized foundation-model benchmarks, unrelated disease contexts, or a broad range of downstream tasks. Moreover, because adaptation occurs in a task-specific projection network while the encoder remains frozen, the resulting model is not a reusable robust foundation-model backbone. \textsc{SPECTRA} uses a label-free InfoNCE objective to train parameter-efficient LoRA modules within the encoder, incorporates both scanner and staining variation, requires no downstream labels, and evaluates robustness and representation retention across PathoROB, HEST, THUNDER, and a 38-task CPTAC benchmark.

\paragraph{Task-agnostic foundation-model robustification.}
Recent work by Filiot et al.\ \citep{filiot2026robustifying} reports a model-agnostic fine-tuning procedure applied to ten pathology foundation models. The study evaluates the resulting encoders with PathoROB, HEST, THUNDER, and Patho-Bench and reports improvements in both robustness and aggregate downstream performance. It released two adapted checkpoints: Phaet, derived from Phikon-v2, and Mascaret, derived from Midnight-12k. However, as of September 2026 the preprint does not include a method section or specify the optimization objective, robustness-training data, construction of training pairs, trainable parameters, or optimization hyperparameters. At the time of writing, no training implementation has been released. Consequently, the reported procedure cannot be independently reproduced, audited, or applied to the other eight evaluated backbones.

\textsc{SPECTRA} addresses this reproducibility and accessibility gap by providing a fully specified, label-free InfoNCE objective, parameter-efficient LoRA adaptation, and an open training pipeline based on the public PLISM matched-acquisition dataset \citep{ochi2024plism}.

\section{Methods}
\label{sec:methods}

\subsection{Overview}

\textsc{SPECTRA} is a model-agnostic, parameter-efficient fine-tuning framework which utilizes registered views of the same tissue acquired under different scanner and staining conditions to improve the robustness of a base pathology foundation model. For each model, we freeze the original backbone and train LoRA adapters on the PLISM dataset (\S\ref{sec:data}--\ref{sec:objective}). Only PLISM supplies gradient updates. HEST, CPTAC, and THUNDER are used exclusively for downstream evaluation; PathoROB is used for stopping and checkpoint selection (\S\ref{sec:selection}).

\subsection{Training data}
\label{sec:data}

We use the publicly available PLISM dataset \citep{ochi2024plism}, although \textsc{SPECTRA} is not specific to this dataset and requires only spatially matched tissue images acquired under different conditions. PLISM was designed to isolate acquisition-related variation using serial sections of a tissue microarray containing 46 human tissue types, stained under 13 distinct H\&E protocols and imaged using seven whole-slide scanners and six smartphone cameras \citep{ochi2024plism}. It comprises two spatially registered subsets: PLISM-wsi, containing 91 whole-slide images representing every combination of the 13 stains and seven scanners, and PLISM-sm, containing corresponding fields of view acquired using both scanners and smartphones. We use only PLISM-wsi in this study.

We define each scanner--stain pair as an acquisition condition
$c=(\text{scanner},\text{stain})$. Rather than extracting tiles directly from the original whole-slide images, we use Owkin's re-registered tile release derived from PLISM-wsi \citep{filiot2025distill}. In this release, each of the 91 whole-slide images is represented by the same ordered set of $N=16{,}278$ spatial locations, extracted as $224\times224$-pixel tiles. We denote the tile from location $i$ under condition $c$ by $x_{c,i}$; consequently, $x_{c,i}$ and $x_{c',i}$ depict corresponding tissue regions under different acquisition conditions. Across scanners at a fixed stain, corresponding tiles originate from the same physical tissue section. Across stains, they originate from adjacent serial sections, which preserve tissue identity and local architecture but do not contain exactly the same cells \citep{ochi2024plism}.

To evaluate generalization beyond the acquisition conditions observed during training, we hold out two scanners (GT450, S210) and three stains (HRH, KR, MY). The remaining five scanners and ten stains define 50 training conditions, while the other 41 conditions contain at least one held-out scanner or stain; the split is fixed across seeds, and tissue locations are shared between training and held-out acquisition conditions (Appendix~\ref{app:evaluation}).

\subsection{Base models}
We apply \textsc{SPECTRA} to a diverse set of pathology foundation models to assess the consistency of its improvements and to compare robustness and downstream performance before and after fine-tuning. We evaluate Phikon-v2 \citep{phikonv2}, Midnight-12k \citep{midnight}, Virchow2 \citep{virchow2}, H-optimus-0 \citep{hoptimus0}, UNI2-h \citep{uni2h,chen2024uni}, Virchow \citep{virchow}, and OpenMidnight 1.1 \citep{openmidnight2026}, a further optimized release of OpenMidnight \citep{openmidnight}.

\subsection{Grid-batch construction}
\label{sec:grid}

\begin{figure}[t]
\centering
\includegraphics[width=\linewidth]{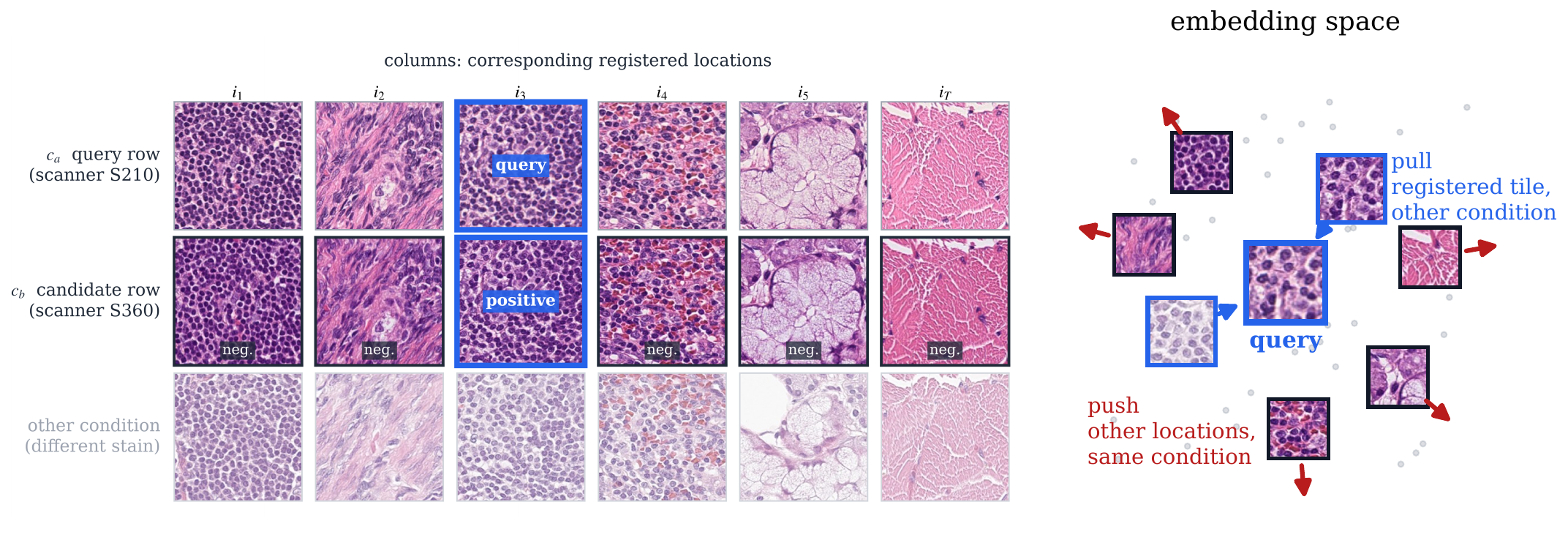}
\caption{Left: schematic grid batch on PLISM tiles; columns are registered locations, rows are conditions. Scanner labels are illustrative; S210 is held out during training. Right: for the query, the registered tile of another condition is pulled in and the other $T-1$ tiles of that condition are pushed away.}
\label{fig:grid}
\end{figure}

At each optimizer step we sample $C$ training conditions and $T$ locations, both without replacement, giving a batch of $CT$ images. For an ordered pair $(c_a,c_b)$, $a\neq b$, and location $i$
(Figure~\ref{fig:grid}), the query is $x_{c_a,i}$, the candidate set is the row of $c_b$ alone; the
positive is $x_{c_b,i}$, and the $T-1$ other tiles in that row are negatives. 

Both orderings are included. All candidates in a softmax share one acquisition condition, so acquisition identity alone cannot identify the matching tissue location. The recipe uses $C=2$,
$T=900$, so for each step we have 1,800 images and 899 negatives per query
(alternative allocations of $C$ and $T$ are ablated in \S\ref{sec:ablations}).

\subsection{CLS and patch-token contrastive objectives}
\label{sec:objective}
Rank-32 LoRA \citep{lora} adapters on the attention and MLP blocks train under $5\%$ of the backbone parameters ($2.7$--$4.4\%$, adapters only). The LoRA-adapted backbone maps a tile $x_{c,i}$ to a CLS token and $P$ patch tokens
$u_1,\dots,u_P$ (register tokens discarded). This gives the two views: $v^{\textsc{cls}}_{c,i}$
is the CLS token, and $v^{\textsc{gem}}_{c,i}$ is a softplus generalized mean (GeM) \citep{gem} of the
patch tokens with learned exponent $\rho$, which controls how strongly large activations contribute (Appendix~\ref{app:pseudo}, Eq.~\eqref{eq:gem}). Each view passes through
its own two-layer MLP head and is $\ell_2$-normalized, giving unit vectors $z^h_{c,i}$ for $h\in\{\textsc{cls},\textsc{gem}\}$.

For an ordered pair of conditions $(c_a,c_b)$, $a\neq b$, query $i$ and candidate $j$, let
$\ell^h_{abij}=\langle z^h_{c_a,i},z^h
_{c_b,j}\rangle/\tau$ be their cosine similarity scaled by temperature $\tau=0.07$. Each view $h$ is trained with the same reweighted InfoNCE loss
\citep{infonce},
\begin{equation}
\mathcal{L}_h=-\frac{1}{C(C-1)T}
\sum_{a\neq b}\sum_{i=1}^{T}
\log\frac{\exp(\ell^h_{abii})}{\sum_{j=1}^{T}w^h_{ij}\exp(\ell^h_{abij})},
\end{equation}
where the weights $w^h_{ij}$ apply pseudo-tissue reweighting~\eqref{eq:w}. Each location $i$ carries a fixed
pseudo-group $q_i$: locations are grouped into the eight-connected spatial components of the tile
lattice (putative tissue-microarray cores), which are then merged to 46 groups by clustering
frozen base-model embeddings of one reference condition, so that a pseudo-group approximates one
tissue type (Appendix~\ref{app:pseudo}). A candidate sharing the query's pseudo-group ($q_j=q_i$,
$j\neq i$) is dropped from the GeM softmax and up-weighted by $e^{3}$ in the CLS softmax, while
all other candidates and the positive keep weight $1$ (ablations in \S\ref{sec:ablations}). The two views are weighted equally,
\begin{equation}
\mathcal{L}=0.5\,\mathcal{L}_{\textsc{cls}}+0.5\,\mathcal{L}_{\textsc{gem}}.
\end{equation}

\subsection{Fine-tuning and checkpoint selection}

Only the adapters, the projection heads, and the GeM exponent are trained; the backbone is
otherwise frozen. The heads and GeM exponent are discarded after training. Evaluation uses the
backbone's original embedding and each benchmark's pooling rule. Training runs 500 steps with
AdamW \citep{adamw} (Appendix~\ref{app:implementation}). Adapters are merged into the base
weights.

\paragraph{Checkpoint selection.}\label{sec:selection}
We perform three independent adapter fine-tuning runs per backbone and select one checkpoint per seed by PathoROB RI
(defined in \S\ref{sec:experiments}) with a tolerance rule ($\epsilon=0.007$, Appendix~\ref{app:selection}). All 21 selected checkpoints fall at steps $100$--$200$, within the 200-step linear warmup.

\section{Experiments}
\label{sec:experiments}

We test whether adaptation against scanner and staining variation improves the usefulness of pathology foundation-model representations across downstream tasks. Our evaluation spans robustness to site and acquisition shifts (PathoROB), cross-condition tissue retrieval (PLISM), and downstream benchmarks, including spatial transcriptomics prediction (HEST), CPTAC classification, and general performance (THUNDER). Full evaluation details are provided in Appendix~\ref{app:evaluation}.

\paragraph{PathoROB.}
We evaluate centre-level robustness using PathoROB across TCGA $2\times2$, Camelyon, and Tolkach-ESCA, and report the robustness index (RI) averaged across the three benchmarks. Let $SO$ denote the number of same-class, other-centre $k$-NN pairs and $OS$ the number of other-class, same-centre pairs, pooled across queries. We compute
\begin{equation}
\mathrm{RI} = \frac{SO}{SO+OS}.
\end{equation}
Higher RI indicates representations that preserve class similarity over centre-specific similarity.

\paragraph{Held-out PLISM retrieval.}
We test cross-acquisition alignment on the 41 scanner--stain conditions excluded from training; tissue locations are shared with the training set. For each query tile, we perform top-1 retrieval of its registered counterpart among 256 candidates. Scanner-pair evaluations vary the scanner while holding stain fixed; stain-pair evaluations vary the stain.

\paragraph{HEST.}
We follow the HEST evaluation protocol \citep{hest}. We fit PCA--Ridge probes for 50 genes across nine tasks and report mean Pearson correlation $r$ across tasks.

\paragraph{CPTAC.}
We evaluate the CPTAC subset of Patho-Bench \citep{pathobench}, reporting macro one-vs-rest AUC across 38 classification tasks.

\paragraph{THUNDER.}
We use the full THUNDER evaluation suite \citep{thunder} across 16 classification and four segmentation datasets. We report $k$-NN and linear-probe macro-F1 performance, few-shot classification, segmentation, calibration via expected calibration error (ECE), and adversarial robustness.

\paragraph{Evaluation protocol.}
For every benchmark, we compare each base model against its fine-tuned counterpart using the same evaluation harness and the PathoROB-selected checkpoint for each adapter seed. Fine-tuned results are reported as mean $\pm$ two sample standard deviations across three adapter seeds, describing variability across runs, while base-model results are a single evaluation (deterministic except segmentation, Appendix~\ref{app:evaluation}). These spreads describe run variability, not patient-level uncertainty or statistical significance. Per-task HEST and per-dataset THUNDER seed spreads are tabulated in Appendix Tables~\ref{tab:hestranks}, \ref{tab:ranks} and~\ref{tab:ranks-b}.

\section{Results}
\label{sec:results}

\begin{figure}[H]
\centering
\includegraphics[width=\linewidth]{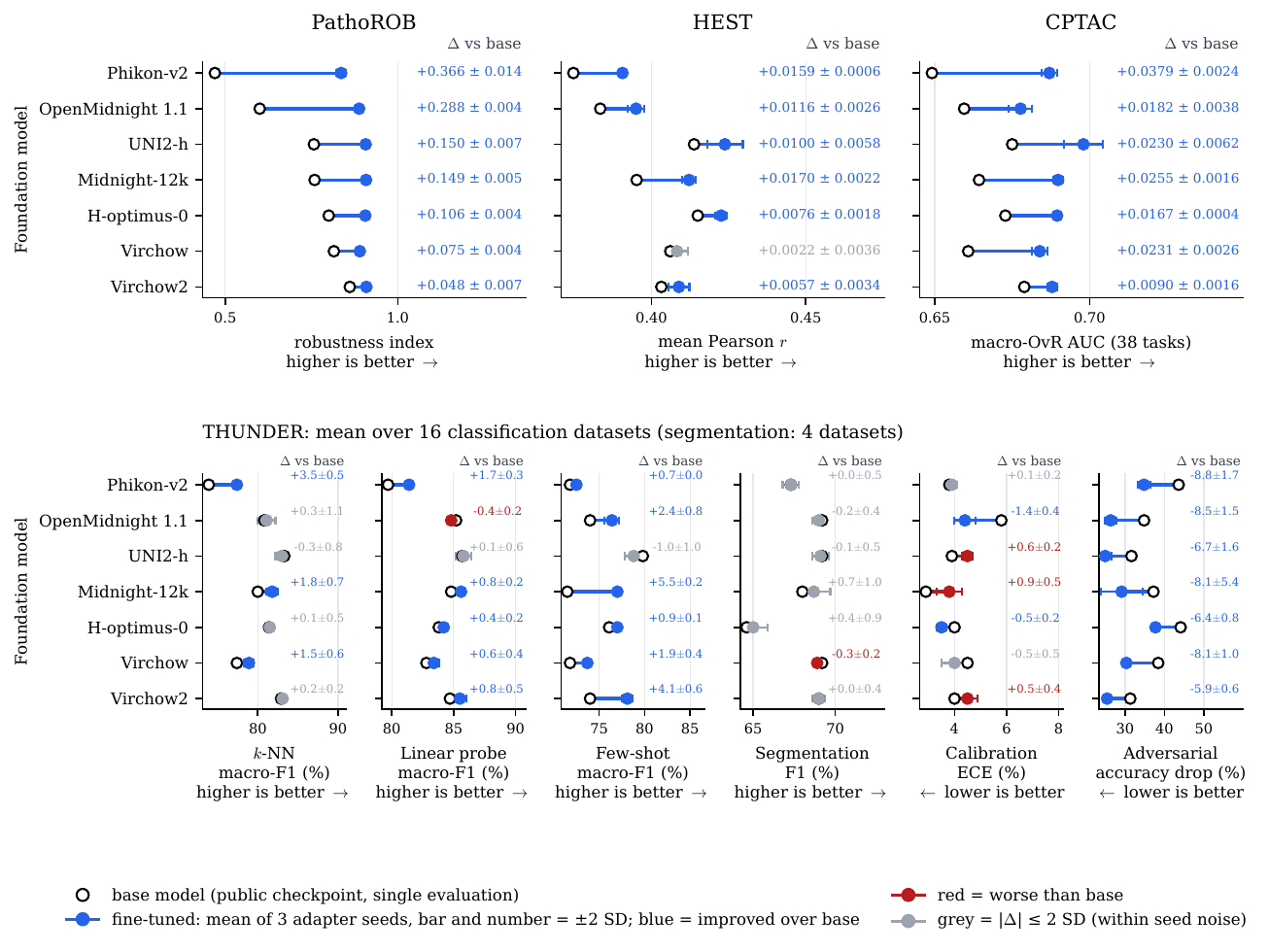}
\caption{Base $\to$ fine-tuned per backbone, ordered by base PathoROB RI. Hollow: base. Filled: fine-tuned mean over three seeds at the selected checkpoint, bar $\pm2$ SD. Numbers are $\Delta$ vs base. Blue improved, red worse, grey: absolute change smaller than $2$ SD.}
\label{fig:b2t}
\end{figure}

Across benchmarks, fine-tuning consistently improves robustness while largely preserving, and in most settings improving, downstream task performance. The clearest effects are on PathoROB and CPTAC, where improvements occur across all backbones. HEST likewise improves on average for every backbone, although sub-tasks have more mixed results. THUNDER's classification and few-shot performance are generally improved, while calibration shows regressions for several models and segmentation differences are largely within evaluation variability. Figure~\ref{fig:b2t} summarizes these base-to-fine-tuned changes, while Table~\ref{tab:overall} places them in the context of each benchmark's published model field.

\begin{table}[htbp]
\caption{Leaderboard position, base $\to$ fine-tuned, in each benchmark's published field:
PathoROB and HEST by per-dataset rank sums, THUNDER by rank sum over six tasks, including the
adversarial-drop task shown in Figure~\ref{fig:b2t} (Appendix Table~\ref{tab:ranks}). Ties share a
position. Each adapted model is ranked separately against the reference field. Fields hold 23, 26 and 32 models. $^\dagger$: absent from the published field and
inserted into it (24 models) rather than substituted (Appendix~\ref{app:evaluation}).}
\label{tab:overall}
\centering
\small
\setlength{\tabcolsep}{4pt}
\renewcommand{\arraystretch}{0.95}
\input{tables/overall_ranks}
\end{table}

\subsection{Cross-centre robustness: PathoROB}

\begin{table}[htbp]
\caption{PathoROB mean RI, RI rank sum and dense-rank position in the 23-model published field,
base $\to$ fine-tuned ($\pm$ two SDs over three seeds).
$^\dagger$: rank estimated by insertion (Appendix~\ref{app:evaluation}).
Last column: held-out PLISM top-1 retrieval (256 candidates, scanner\,/\,stain; two SDs $\leq0.036$); its held-out axis is acquisition, not tissue.}
\label{tab:pathorobranks}
\centering
\small
\setlength{\tabcolsep}{4pt}
\renewcommand{\arraystretch}{0.95}
\input{tables/pathorob_ranks}
\end{table}

Fine-tuning improves cross-centre robustness for every backbone. Mean RI increases in all cases, by $+0.048$ to $+0.366$, each exceeding the fine-tuned $\pm2$\,SD spread (Figure~\ref{fig:b2t}, Table~\ref{tab:pathorobranks}), with the largest per-dataset gains occurring on Camelyon (Appendix Table~\ref{tab:pathorob-sub}). Across models, fine-tuned RI values also tend to converge toward approximately $0.9$, suggesting diminishing gains among backbones that already begin with relatively strong centre-robust representations.

These improvements in RI do not correspond to a substantial loss in predictive structure. PathoROB's prediction-performance submetric, pooled $k$-NN class purity, decreases on TCGA for every backbone by $0.010$--$0.030$, while $k$-NN balanced accuracy averaged over the three datasets changes by only $-0.005$ to $+0.003$ (on TCGA alone, $-0.015$ to $+0.009$; Appendix Table~\ref{tab:pathorob-sub}). Thus, the large changes in robustness are accompanied by comparatively small changes in local class-discriminative performance.

\subsection{Held-out PLISM retrieval}

Top-1 retrieval of the registered counterpart improves for every backbone on both axes
(Table~\ref{tab:pathorobranks}): $0.752$--$0.978$ to $0.991$--$0.997$ across scanners, and
$0.560$--$0.830$ to $0.883$--$0.931$ across stains. Stain remains the harder axis both before
and after (\S\ref{sec:data}).

\subsection{HEST and CPTAC}

\begin{table}[htbp]
\caption{HEST/CPTAC performance: base $\to$ fine-tuned. HEST is mean Pearson $r$ over nine tasks with the
rank sum over those tasks and the positional rank in the 27-model published field; CPTAC is
macro-OvR AUC over 38 Patho-Bench classification tasks. Per-task results are in
Appendix Tables~\ref{tab:hestranks}, \ref{tab:cptac-per-task} for HEST and CPTAC, respectively.}
\label{tab:hestcptac}
\centering
\small
\setlength{\tabcolsep}{4pt}
\renewcommand{\arraystretch}{0.95}
\input{tables/hest_cptac}
\end{table}

The robustness gain is accompanied by gains on both downstream benchmarks. Mean HEST Pearson $r$ increases for every backbone, by $+0.002$ to $+0.017$, while CPTAC macro-OvR AUC increases for every backbone, by $+0.009$ to $+0.038$ (Figure~\ref{fig:b2t}, Appendix Table~\ref{tab:hestranks}, Appendix Table~\ref{tab:cptac-per-task}).

The HEST improvements are not uniform across individual gene-prediction tasks: some backbone-task pairs improve while others regress. The mean change nonetheless exceeds the fine-tuned $\pm2$\,SD spread on six of the seven; on Virchow ($+0.002$ against $\pm0.004$) it lies within seed spread. Fine-tuning therefore does not degrade, and on most backbones modestly improves, mean transcriptomic prediction. Against the per-task seed spread (Appendix Table~\ref{tab:hestranks}), 33 of the 42 per-task improvements and 9 of the 21 regressions exceed $\pm2$\,SD. All seven CPTAC gains exceed the fine-tuned $\pm2$\,SD spread; the smallest is Virchow2 ($+0.009$ against $\pm0.002$) and the widest spread is UNI2-h ($+0.023$ against $\pm0.006$).

\subsection{THUNDER}

THUNDER provides a broader and less uniformly positive picture than PathoROB, HEST, or CPTAC; aggregate task means and ranks are in Appendix Table~\ref{tab:ranks}, per-dataset changes in Appendix Figure~\ref{fig:thunderds}. $k$-NN macro-F1 improves on six of seven backbones ($-0.3$ to $+3.5$ points) and linear-probe macro-F1 on six of seven ($-0.4$ to $+1.7$). Few-shot performance varies the most, from $-1.0$ to $+5.5$ points, and the only decrease, UNI2-h ($-1.0$), equals its $\pm1.0$ seed spread (per-dataset spreads in Appendix Table~\ref{tab:ranks}). The rank sum improves for every backbone except UNI2-h, whose sum worsens from 17 to 23 while it keeps first position. Segmentation differences are comparatively small and lie within the repeat variability of the THUNDER harness (Appendix~\ref{app:evaluation}); because segmentation evaluation is non-deterministic, our measured base values can themselves differ slightly from those reported by THUNDER.

Calibration is the main metric that trends negatively. Mean ECE increases for four of seven backbones, beyond the $\pm2$\,SD seed spread on Midnight-12k, Virchow2 and UNI2-h; the degradation is concentrated on BRACS, which worsens on five of seven and beyond $\pm2$\,SD on Midnight-12k, Virchow2 and Virchow (Appendix Figure~\ref{fig:thunderds}, Appendix Table~\ref{tab:ranks-b}). These results show that improved discrimination does not guarantee improved calibration.

The most consistent classification regression occurs on TCGA-uniform, which decreases in 20 of the 21 cells formed by the seven backbones and the $k$-NN, linear-probe, and few-shot tasks, by $5.0$ points on average over those 20 cells. Together with the TCGA purity decrease in PathoROB, this is compatible with reduced shortcut use or loss of useful biological information (\S\ref{sec:discussion}).

\section{Ablations}
\label{sec:ablations}

We test pseudo-tissue reweighting, batch composition, and adapter
rank with one seed per setting: reweighting on Phikon-v2, Midnight-12k, Virchow2, H-optimus-0, and
UNI2-h, the other two on Phikon-v2 (rank also on Virchow2); values are in
Appendix~\ref{app:ablations}, Appendix Table~\ref{tab:ablations}. Reweighting
(Appendix~\ref{app:pseudo}) versus none, at the reweighted arm's selected checkpoint, moves RI by
$+0.009$ on Phikon-v2 and by less than the $\epsilon=0.007$ selection tolerance on the other four,
so we do not claim it as a general mechanism; it also reduces THUNDER classification scores in
14 of 15 backbone-by-probe comparisons (Appendix~\ref{app:ablations}). Batch composition
(\S\ref{sec:grid}; 750 steps, $CT=3600$) favours few conditions and many locations: RI at step 750
rises monotonically from $0.802$ to $0.824$ as $C$ falls from 24 to 2. Adapter rank
(\S\ref{sec:objective}) shows no monotone trend over $r\in\{8,\ldots,128\}$; $r=128$ is lowest on
Virchow2. No sweep moves RI by more than $0.03$, smaller than the recipe's own
$0.048$--$0.366$ gain, but these single-seed sweeps do not establish the contribution of each
component.

\section{Discussion}
\label{sec:discussion}

These results suggest that matched-acquisition adaptation can improve the usefulness of existing pathology encoders: cross-centre robustness and held-out
cross-acquisition retrieval improve on every backbone, both downstream benchmarks move with them,
and THUNDER, the only mixed benchmark, mixes by task rather than by backbone.

Two properties of the procedure are worth noting. First, every selected checkpoint lands at steps
$100$--$200$ (\S\ref{sec:selection}). The robustness gain therefore comes from an early
update to the adapters, made while the learning rate is still ramping to its peak; running the full 500-step schedule is not what produces it. Second, because PathoROB is the selection criterion, its improvement is not an independent
finding, whereas HEST, CPTAC, and THUNDER measure
what the RI-selected checkpoints do to downstream utility.

The regressions on TCGA-derived tasks (\S\ref{sec:results}) could reflect reduced use of acquisition shortcuts, but could also indicate loss of biologically useful information. The current experiments do not distinguish these explanations. Whether that trade is acceptable depends on the deployment setting, and it is one reason we report the per-task tables rather than only the aggregate ranks.

\section{Limitations}\label{sec:limitations}

\textsc{SPECTRA} is trained using a single tissue-microarray dataset acquired at one institution. Although PLISM provides unusually well-controlled scanner and staining variation, its specific contribution to \textsc{SPECTRA}'s performance remains undetermined because we do not compare training on alternative matched-acquisition datasets. Consequently, we cannot establish whether PLISM is optimal or whether larger, more biologically diverse, or multi-institutional datasets would produce greater robustness and downstream transfer. The observed improvements may also depend on PLISM-specific tissue composition, staining protocols, and scanner coverage.

Registration across stains introduces an additional limitation: stain-matched pairs are drawn from adjacent serial sections (\S\ref{sec:data}), which may exhibit genuine local biological differences. Treating these regions as positive pairs therefore introduces correspondence noise and may encourage the model to suppress biologically meaningful variation. More precise cross-section registration, explicit uncertainty modeling, or objectives that distinguish exact from approximate correspondences could reduce this risk.

Three further caveats apply. All of the evidence measures representation consistency across centres and probe performance, not diagnostic accuracy. All ablations are single seed (\S\ref{sec:ablations}), and several of the reported differences are smaller than the selection tolerance $\epsilon$ (\S\ref{sec:selection}), so the pseudo-tissue reweighting, batch-composition, and adapter-rank trends should be read as indicative rather than established; LoRA versus full fine-tuning, one versus two views, temperature, and the number of training conditions are untested. PathoROB is used for checkpoint selection, so its reported gains are optimistically biased relative to the other benchmarks. Finally, the improvements are not uniform: $k$-NN class purity on TCGA, TCGA-uniform classification, and calibration on several backbones regress, and users should check the per-task tables for the tasks they care about before substituting a fine-tuned checkpoint.

\section{Conclusion}

A useful pathology foundation model should capture tissue biology reliably across sites, staining protocols, and scanners. This study shows that existing encoders can move towards that goal with a small trainable update using publicly available images, while simultaneously improving performance on most downstream tasks. Matched acquisitions can therefore serve as a practical resource for improving foundation models after pretraining. We release our training and evaluation codebases, along with adapters and model weights for selected backbones so that our procedure can be reproduced, audited, and applied to other backbones.

\bibliographystyle{plainnat}
\bibliography{references}

\input{appendix}
\clearpage

\end{document}

%% file: tables/overall_ranks.tex
\begin{tabular}{lccc}
\toprule
Backbone & PathoROB\,$\downarrow$ & HEST\,$\downarrow$ & THUNDER\,$\downarrow$ \\
\midrule
Phikon-v2 & 21$\to$\gain{6} & 15$\to$\gain{10} & 18$\to$\gain{14} \\
Midnight-12k & 10$\to$\gain{2$^\dagger$} & 8$\to$\gain{5} & 5$\to$\gain{2} \\
Virchow2 & 3$\to$\gain{2} & 6$\to$\gain{5} & 3$\to$\gain{2} \\
H-optimus-0 & 9$\to$\gain{2} & 3$\to$\gain{2} & 8$\to$\gain{5} \\
UNI2-h & 10$\to$\gain{2} & 4$\to$\gain{1} & 1$\to$1 \\
Virchow & 6$\to$\gain{2} & 5$\to$5 & 11$\to$\gain{6} \\
OpenMidnight 1.1 & 15$\to$\gain{2$^\dagger$} & 13$\to$\gain{9$^\dagger$} & 7$\to$\gain{3$^\dagger$} \\
\bottomrule
\end{tabular}

%% file: tables/pathorob_ranks.tex
\begin{tabular}{lcccc}
\toprule
 & \multicolumn{1}{c}{PathoROB (cross-centre)} & & \multicolumn{1}{c}{PLISM (cross-acquisition)} \\
\cmidrule(lr){2-2}\cmidrule(lr){4-4}
Backbone & Mean RI\,$\uparrow$ & $\Sigma$rank\,(pos.)\,$\downarrow$ & top-1 scanner / stain\,$\uparrow$ \\
\midrule
Phikon-v2 & 0.470$\to$\gain{0.836$\pm$0.014} & 58~(21)$\to$\gain{22~(6)} & 0.858$\to$\gain{0.995$\pm$0.001} / 0.696$\to$\gain{0.897$\pm$0.003} \\
Midnight-12k & 0.759$\to$\gain{0.908$\pm$0.005} & 21~(10)$\to$\gain{5~(2)}$^\dagger$ & 0.752$\to$\gain{0.991$\pm$0.005} / 0.560$\to$\gain{0.883$\pm$0.036} \\
Virchow2 & 0.861$\to$\gain{0.909$\pm$0.007} & 13~(3)$\to$\gain{6~(2)} & 0.854$\to$\gain{0.991$\pm$0.006} / 0.696$\to$\gain{0.903$\pm$0.025} \\
H-optimus-0 & 0.800$\to$\gain{0.906$\pm$0.004} & 27~(9)$\to$\gain{8~(2)} & 0.978$\to$\gain{0.995$\pm$0.001} / 0.830$\to$\gain{0.915$\pm$0.002} \\
UNI2-h & 0.757$\to$\gain{0.907$\pm$0.007} & 30~(10)$\to$\gain{6~(2)} & 0.939$\to$\gain{0.997$\pm$0.000} / 0.761$\to$\gain{0.913$\pm$0.031} \\
Virchow & 0.815$\to$\gain{0.890$\pm$0.004} & 25~(6)$\to$\gain{14~(2)} & 0.758$\to$\gain{0.997$\pm$0.000} / 0.597$\to$\gain{0.931$\pm$0.006} \\
OpenMidnight 1.1 & 0.600$\to$\gain{0.888$\pm$0.004} & 42~(15)$\to$\gain{8~(2)}$^\dagger$ & 0.868$\to$\gain{0.995$\pm$0.002} / 0.682$\to$\gain{0.912$\pm$0.013} \\
\bottomrule
\end{tabular}

%% file: tables/hest_cptac.tex
\begin{tabular}{lccc}
\toprule
Backbone & HEST mean $r$\,$\uparrow$ & HEST $\Sigma$rank\,(position)\,$\downarrow$ & CPTAC AUC\,$\uparrow$ \\
\midrule
Phikon-v2 & 0.3747$\to$\gain{0.3906$\pm$0.0006} & 131~(15)$\to$\gain{97~(10)} & 0.6491$\to$\gain{0.6870$\pm$0.0024} \\
Midnight-12k & 0.3952$\to$\gain{0.4122$\pm$0.0022} & 97~(8)$\to$\gain{51~(5)} & 0.6643$\to$\gain{0.6898$\pm$0.0016} \\
Virchow2 & 0.4032$\to$\gain{0.4089$\pm$0.0034} & 65~(6)$\to$\gain{48~(5)} & 0.6789$\to$\gain{0.6879$\pm$0.0016} \\
H-optimus-0 & 0.4150$\to$\gain{0.4226$\pm$0.0018} & 37~(3)$\to$\gain{29~(2)} & 0.6728$\to$\gain{0.6895$\pm$0.0004} \\
UNI2-h & 0.4138$\to$\gain{0.4238$\pm$0.0058} & 42~(4)$\to$\gain{24~(1)} & 0.6750$\to$\gain{0.6980$\pm$0.0062} \\
Virchow & 0.4061$\to$0.4083$\pm$0.0036 & 67~(5)$\to$\gain{56~(5)} & 0.6608$\to$\gain{0.6839$\pm$0.0026} \\
OpenMidnight 1.1 & 0.3834$\to$\gain{0.3950$\pm$0.0026} & 119~(13)$\to$\gain{88~(9)} & 0.6595$\to$\gain{0.6777$\pm$0.0038} \\
\bottomrule
\end{tabular}

%% file: appendix.tex
\appendix

\section{Method details}
\label{app:pseudo}

\paragraph{GeM view.} With $P$ patch tokens $u_1,\dots,u_P$ and learned exponent
$\rho\in[1,8]$, the \textsc{gem} view is
\begin{equation}
v^{\textsc{gem}}_{c,i}
=\Big(\frac{1}{P}\sum_{p=1}^{P}\operatorname{softplus}(u_p)^{\rho}\Big)^{1/\rho},
\label{eq:gem}
\end{equation}
softplus and both powers in~\eqref{eq:gem} acting elementwise.

\paragraph{Reweighting.} Each location $i$ carries a fixed pseudo-group $q_i$, meant to represent
the same tissue type. The softmax weight of candidate $j$ against query $i$ for head $h$ is
\begin{equation}
w^h_{ij}=\begin{cases}
\omega_h,&q_i=q_j\ \text{and}\ i\neq j,\\
1,&\text{otherwise,}
\end{cases}
\label{eq:w}
\end{equation}
with $\omega_{\textsc{cls}}=e^{3}\approx20$ and $\omega_{\textsc{gem}}=0$. The positive has
$w^h_{ii}=1$ and is never reweighted; $\omega_h=1$ in~\eqref{eq:w} recovers standard InfoNCE. This up-weights same-pseudo-group negatives in the CLS loss and excludes them from the GeM loss.

\paragraph{Map construction.} We construct the same pseudo-groups for all models before training. First, the row and column stored in every tile
key are placed on a binary occupancy lattice. 
Directionally connected components along all eight directions are merged to create contiguous spatial regions. 
A component smaller than one quarter of the
median component size is assigned to the nearest larger component by centroid distance. This
geometry-only stage results in 49 groupings, which is 3 more than the 46 tissue types.

We then embed every tile from one reference condition (GMH stain, S60 scanner) with frozen
Virchow2, concatenating its CLS and mean-patch embeddings. Embeddings are averaged and
$\ell_2$-normalized within each spatial component, and cosine distances are computed between
the 49 component means. Average-linkage hierarchical clustering is cut at 46 groups, using
PLISM's documented category count as the intended grouping and thereby forcing three
merges. This does not guarantee that the recovered components correspond one-to-one
with the 46 categories. Every coordinate inherits its component's cluster.

\section{Checkpoint selection}
\label{app:selection}

\begin{figure}[htbp]
\centering
\includegraphics[width=\linewidth]{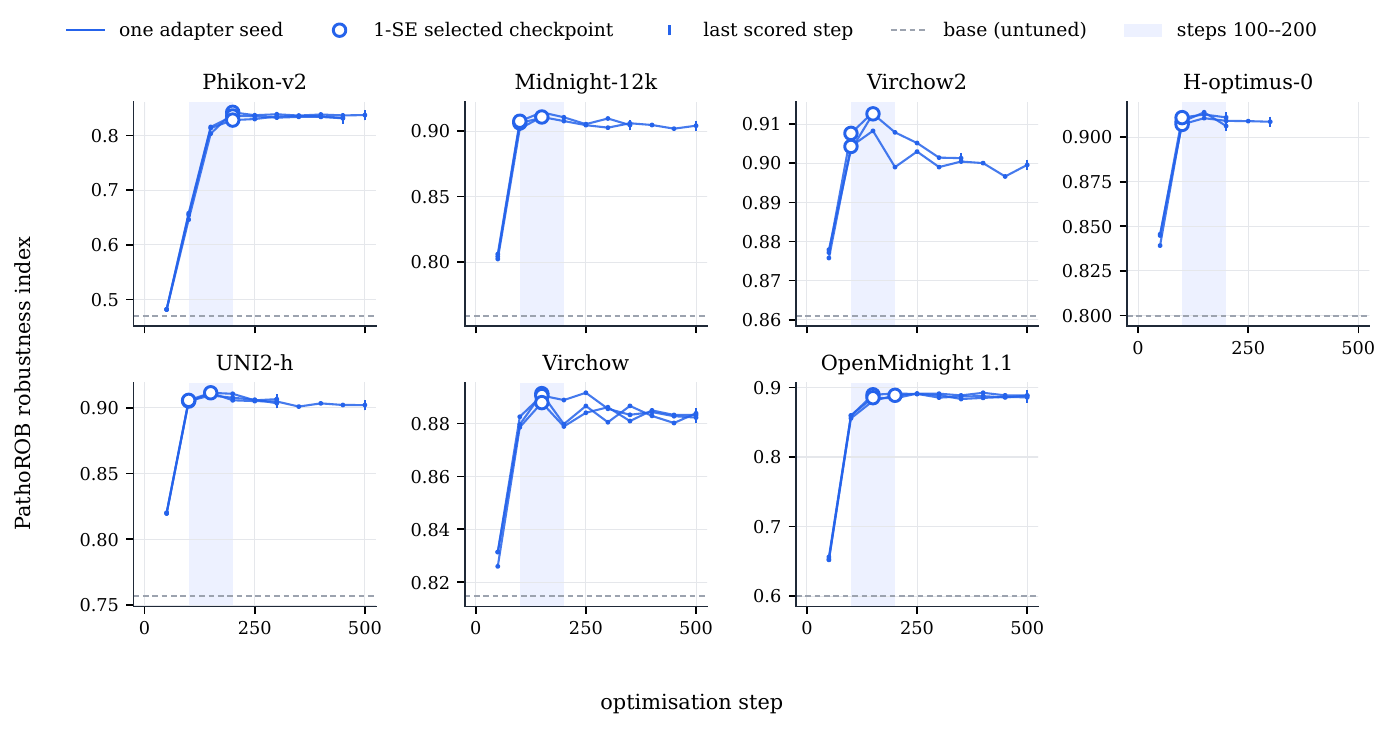}
\caption{PathoROB RI against step for all 21 runs (three seeds per backbone). Circles mark
the selected checkpoint, the band is steps $100$--$200$, and the dashed line is the base
model. Each panel has its own $y$-range.}
\label{fig:ristep}
\end{figure}

Let $R_t$ be PathoROB average RI at checkpoint index $t$ (evaluated every 50 optimizer steps) and $B_t=\max_{s\le t}R_s$. Per seed,
\begin{align}
t_{\mathrm{stop}} &= \min\{\,t\ge 2 : R_t-B_{t-1}\le\epsilon\,\}, \\
t^\star &= \min\{\,t\le t_{\mathrm{stop}} : R_t\ge B_{t_{\mathrm{stop}}}-\epsilon\,\},
\end{align}
with $\epsilon=0.007$, the largest average-RI difference in a two-run pilot. Every selected step
lies in $100$--$200$ (Figure~\ref{fig:ristep}, Table~\ref{tab:selected}), inside the 200-step
warmup and under one pass over the training grid.

\begin{table}[htbp]
\caption{Selected checkpoint step per backbone and adapter seed under the rule above.
Columns are seed identifiers: Midnight-12k, Virchow2 and H-optimus-0 use seeds 0, 1, 3, the
others 0, 1, 2.}
\label{tab:selected}
\centering
\begin{tabular}{lccc}
\toprule
Backbone & Seed 0 & Seed 1 & Seed 2\,/\,3 \\
\midrule
Phikon-v2 & 200 & 200 & 200 \\
Midnight-12k & 150 & 100 & 100 \\
Virchow2 & 100 & 150 & 100 \\
H-optimus-0 & 100 & 100 & 100 \\
UNI2-h & 100 & 150 & 100 \\
Virchow & 150 & 150 & 150 \\
OpenMidnight 1.1 & 200 & 150 & 150 \\
\bottomrule
\end{tabular}
\end{table}

\section{Evaluation protocol details}
\label{app:evaluation}

\paragraph{PathoROB.}
We run the benchmark's own harness unchanged at a pinned revision, so RI follows its published
definition~\citep{koemen2026pathorob} (\S\ref{sec:experiments}): for every query tile the $k$ nearest neighbours are found
by cosine similarity over $\ell_2$-normalized embeddings. Embeddings are the CLS token concatenated with the mean patch token
for every backbone, the pooling of the benchmark's reference rows, without projection heads.
$k=61$, $11$, $46$ for TCGA $2\times2$, Camelyon, Tolkach-ESCA, the harness's per-dataset defaults. \texttt{prediction\_performance}
is the fraction of $k$ nearest neighbours sharing the biological class, pooled over same- and
other-centre neighbours. The submetrics tabulated in Appendix Table~\ref{tab:pathorob-sub}
follow the same harness: balanced accuracy is $k$-NN balanced accuracy at $k_{\mathrm{opt}}$ (the neighbourhood size selected by the benchmark),
and ID, OOD and pooled purity are the fractions of same-centre, other-centre and all $k$
nearest neighbours that share the biological class. All five are higher-is-better. Our base RI reproduces the published rows to $0.004$ or better except
H-optimus-0 (TCGA $0.803$ vs $0.812$, Camelyon $0.679$ vs $0.705$); both H-optimus-0 columns are
our own measurement. Midnight-12k, and OpenMidnight 1.1 are absent from the published field, so they
are ranked by insertion rather than substitution, making that field 24 models rather than 23.

\paragraph{Held-out PLISM retrieval.}
Every condition is sampled at the same 256 coordinates. The embedding concatenates the CLS token and the mean patch token, with no projection head. A pair is two acquisition conditions that differ in scanner alone or in stain alone, and both conditions are held out from training. One supplies the 256 query tiles, the other the 256 candidates at the same coordinates. We rank candidates by cosine similarity after $\ell_2$ normalisation and score top-1 accuracy per pair, then average over pairs. PLISM-wsi spans seven scanners (AT2, GT450, P, S210, S360, S60, SQ) and 13 staining protocols (GIV, GIVH, GM, GMH, GV, GVH, HR, HRH, KR, KRH, LM, LMH, MY, in PLISM's codes). The 41 held-out conditions (\S\ref{sec:data}) comprise 20 with an unseen scanner, 15 with an unseen stain, and six with both. This gives 73 cross-scanner pairs and 171 cross-stain pairs. These aggregates test held-out acquisition conditions; the varied scanner or stain is not necessarily unseen.

\paragraph{HEST.}
We use the official train/test splits and 50-gene list per task. Standardization and
256-dimensional PCA are fit on training embeddings only. Ridge is applied with HEST defaults. We report Pearson $r$ per
gene, averaged over genes within a task, then over nine tasks. Pooling is CLS for Phikon-v2,
Midnight-12k, H-optimus-0, UNI2-h, OpenMidnight, OpenMidnight 1.1, and concatenated CLS and mean-patch for
Virchow2 and Virchow, identical for base and fine-tuned. Base averages reproduce published rows
to $0.0010$.

\paragraph{CPTAC.}
We test models across 38 Patho-Bench \citep{pathobench} CPTAC classification tasks, on the benchmark's published splits: nine on
COAD, five on LUAD, four each on CCRCC and LSCC, three each on BRCA, GBM, HNSC and UCEC, two on
PDA, and one each on LUNG (subtype) and OV. Targets are gene-mutation status, MSI-H, histologic
grade, immune class and subtype. Tile embeddings are mean-pooled to a slide vector and, for
case-level tasks, slides are mean-pooled to a case vector; the probe is the benchmark's
class-balanced logistic regression ($C=0.5$, 100 bootstraps), and the metric is macro
one-vs-rest AUC, averaged without weighting over the 38 tasks (Table~\ref{tab:hestcptac}; per
task in Table~\ref{tab:cptac-per-task}). Pooling is CLS for Phikon-v2, H-optimus-0, UNI2-h, OpenMidnight 1.1, and concatenated CLS and mean-patch for
Virchow2, Virchow, and Midnight-12k, identical for base and fine-tuned. The benchmark's four survival tasks are not scored.

\paragraph{THUNDER.}
Both terms of every base $\to$ fine-tuned pair are our own measurement under the benchmark's
harness, and a rank is obtained by substituting that value for the backbone's published row in
the 32-model field, so base and fine-tuned ranks are comparable to each other and to the
published field. The segmentation probe is non-deterministic, so a segmentation spread folds in
harness repeat variability as well as adapter-seed variability, and our measured base can differ
slightly from the published one. ECE is the linear-probe expected calibration error, estimated
from relatively few populated bins, which makes small changes sensitive to how predictions fall
across bins. ECE and the adversarial macro-F1 drop are lower-is-better; every other THUNDER task
is higher-is-better.

\paragraph{Outliers and consistent effects.}
Three patterns dominate the per-dataset panel. TCGA-uniform is the only dataset with a
unanimous macro-F1 regression across two tasks: $k$-NN falls on all seven backbones ($-10.9$ to $-2.5$, mean $-6.6$, every cell beyond its own two SDs) and linear probing on all seven ($-6.6$ to $-0.9$, five beyond two SDs). Of the 21 cells spanned by $k$-NN, linear probing and few-shot, 20 decrease, and those 20 average $5.0$ points. Adversarial
robustness, on the contrary, has the most consistent positive effect in the panel: the F1 drop shrinks on all seven backbones for 9 of the 16 datasets and worsens unanimously on none, with PCam the extreme ($-27.0$ to $-12.9$, mean $-18.4$, all seven beyond two SDs) followed by CCRCC ($-29.5$ to $-6.4$, mean $-13.8$) and BACH ($-21.1$ to $-7.3$, mean $-11.2$). The ECE degradation is concentrated on the two small breast datasets,
which supply the two largest ECE increases exceeding two SDs: BRACS worsens on five of seven backbones ($+2.3$ to $+13.5$, three beyond two SDs, the exceptions being H-optimus-0 and OpenMidnight 1.1), and BACH on UNI2-h
moves $8.4\to23.4$, the largest in the panel, while its BACH macro-F1 rises $2.4$. Excluding
BRACS, the per-backbone mean $\Delta$ECE over the remaining 15 datasets is $-0.20$, $+0.82$,
$-0.39$, $-0.41$, $+0.45$, $-0.85$, $-1.23$, so only Midnight-12k and UNI2-h still rise.
Segmentation shows no systematic pattern on any of its four datasets. The volatile cells are the
small ones: BACH has 132 test images over four classes with 23 in the smallest, and BRACS 570
over seven, against $32{,}768$ for PCam and $56{,}275$ for TCGA-TILS. TCGA-uniform is large
($56{,}200$) but 32-way, with 180 in its smallest class. Published THUNDER mean ECE spans
$2.9$--$5.8$ across pathology encoders.

\section{Full result tables}
\renewcommand{\floatpagefraction}{0.5}

\begin{figure}[t]
\centering
\includegraphics[width=\linewidth]{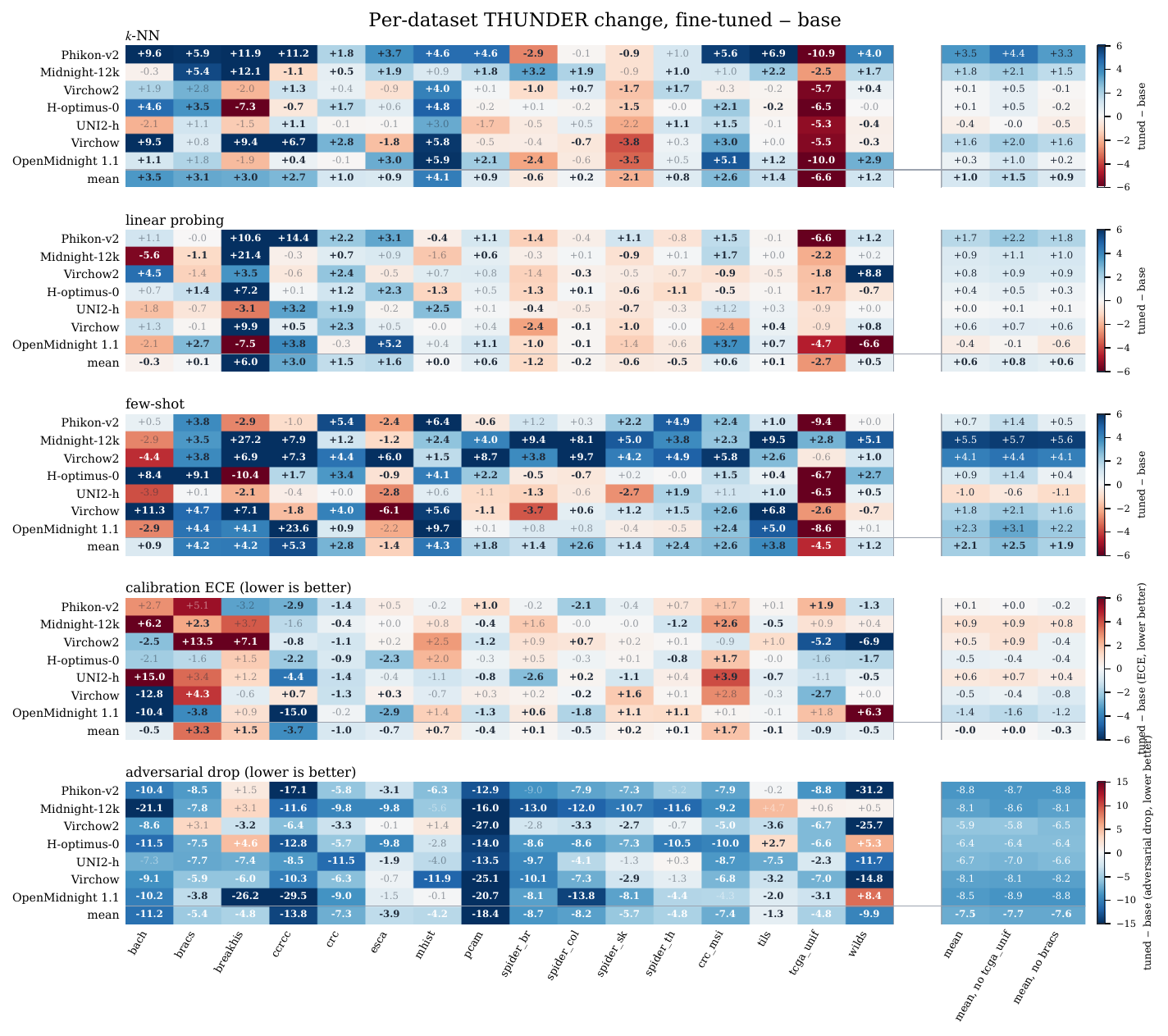}
\caption{THUNDER. Per-dataset change, fine-tuned $-$ base; bold exceeds two SDs on that
dataset, faint lies within. In the ECE and adversarial panels (lower is better) the colour
scale is reversed, so blue always improves; the adversarial scale is wider. Bottom row of each
panel: mean across backbones. Right-hand columns: row means, the last two dropping
tcga\_uniform and BRACS.}
\label{fig:thunderds}
\end{figure}

We report full tables per benchmark, in the order of the main text: PathoROB (Table~\ref{tab:pathorob-sub}; RI rank sums and PLISM retrieval are in main-text Table~\ref{tab:pathorobranks}), HEST (Table~\ref{tab:hestranks}), CPTAC (Table~\ref{tab:cptac-per-task}), and THUNDER (Table~\ref{tab:ranks}, continued in Table~\ref{tab:ranks-b}; Figure~\ref{fig:thunderds}).
Every fine-tuned value in the paper is a mean over three adapter seeds at each seed's selected checkpoint (Table~\ref{tab:selected}). Base values are single evaluations (deterministic except segmentation). Per-seed summary blocks list the unaggregated values behind the aggregate rows: the base value and each adapter seed's value at its own selected checkpoint, seed columns labelled a--c and ordered by seed identifier within a backbone (0, 1, 2, or 0, 1, 3 for Midnight-12k, Virchow2 and H-optimus-0), so they do not pair the same seed across backbones. Ablations (Appendix~\ref{app:ablations}) are single seed and carry none.
In every value cell, the fine-tuned number is
\gain{bold} when the change from base is an improvement exceeding two sample SDs, \loss{red}
when it is a regression exceeding two sample SDs, and plain when it lies within two SDs. Rank cells have no spread, so a rank is \gain{bold} when it improves and \loss{red} when it worsens.

On HEST (\S\ref{sec:results}), the largest resolved regression is H-optimus-0 on COAD ($-0.022$ against $\pm0.004$). On THUNDER, the fraction of per-dataset cells whose change exceeds two SDs is $71/112$ for $k$-NN, $63/112$ for linear probing, $88/112$ for few-shot, $8/28$ for segmentation, $56/112$ for ECE and $88/112$ for adversarial drop. Of the five BRACS ECE increases, three (Midnight-12k, Virchow2, Virchow) exceed two SDs and two (Phikon-v2, UNI2-h) lie within it.

\begin{table}[p]
\caption{PathoROB. Top: submetrics per dataset, base $\to$ fine-tuned mean $\pm$ two sample
SDs; all five are higher-is-better and are defined in Appendix~\ref{app:evaluation}.
Bottom: per-seed mean RI, base and three adapter seeds.}
\label{tab:pathorob-sub}
\centering
\footnotesize
\setlength{\tabcolsep}{3pt}
\resizebox{\textwidth}{!}{%
\input{tables/pathorob_submetrics}}
\par\medskip
{\footnotesize\emph{Per-seed mean RI}}\par\smallskip
\input{tables/perseed_pathorob}
\end{table}

\begin{table}[p]
\caption{HEST. Top: per task, base $\to$ fine-tuned mean Pearson $r$ $\pm$ two sample SDs on
the upper line and that value's rank in the 27-model published field below it; the Mean $r$ and
$\Sigma$rank columns repeat Table~\ref{tab:hestcptac}.
Bottom: per-seed mean $r$ over the nine tasks.}
\label{tab:hestranks}
\centering
\footnotesize
\setlength{\tabcolsep}{2.5pt}
\resizebox{\textwidth}{!}{%
\input{tables/hest_per_task}}
\par\medskip
{\footnotesize\emph{Per-seed mean $r$}}\par\smallskip
\input{tables/perseed_hest}
\end{table}

\begin{table}[p]
\caption{CPTAC. Top: per task, base $\to$ fine-tuned macro one-vs-rest AUC. Highlighting is
gated on the two-SD spread, which this table does not print; the per-seed block below carries
the values behind it. The Mean row matches Table~\ref{tab:hestcptac}.
Bottom: per-seed macro-OvR AUC.}
\label{tab:cptac-per-task}
\centering
\footnotesize
\setlength{\tabcolsep}{3pt}
\resizebox{\textwidth}{!}{%
\input{tables/cptac_per_task}}
\par\medskip
{\footnotesize\emph{Per-seed macro-OvR AUC}}\par\smallskip
\input{tables/perseed_cptac}
\end{table}

\begin{table}[p]
\caption{THUNDER, part one of two. Top: task means, base $\to$ fine-tuned mean $\pm$ two
sample SDs, each value's rank in the 33-model field in parentheses, then the rank sum and its
implied position; ECE and the adversarial drop are lower-is-better.
Bottom: per-dataset macro-F1 for $k$-NN, linear probing and 16-shot, same spread.
Continued in Table~\ref{tab:ranks-b}.}
\label{tab:ranks}
\centering
\footnotesize
\setlength{\tabcolsep}{3pt}
\resizebox{\textwidth}{!}{%
\input{tables/thunder_ranks_full}}
\par\medskip
{\scriptsize
\setlength{\tabcolsep}{3pt}
\renewcommand{\arraystretch}{0.95}
\resizebox{\textwidth}{!}{%
\input{tables/thunder_per_ds_cls}}}
\end{table}

\begin{table}[p]
\caption{THUNDER, part two of two (continued from Table~\ref{tab:ranks}). Top: per dataset,
segmentation F1, linear-probe ECE and adversarial macro-F1 drop (percentage points, the latter
two lower-is-better; fp32-attack cells), base $\to$ fine-tuned mean $\pm$ two sample SDs; the
segmentation spread also folds in harness repeat variability (Appendix~\ref{app:evaluation}).
Bottom: per-seed task means, percentage points.}
\label{tab:ranks-b}
\centering
{\scriptsize
\setlength{\tabcolsep}{3pt}
\renewcommand{\arraystretch}{0.95}
\resizebox{\textwidth}{!}{%
\input{tables/thunder_per_ds_other}}}
\par\medskip
{\footnotesize\emph{Per-seed task means}}\par\smallskip
\scriptsize
\setlength{\tabcolsep}{3pt}
\resizebox{\textwidth}{!}{%
\input{tables/perseed_thunder}}
\end{table}

\section{Ablation details}
\label{app:ablations}

\paragraph{Pseudo-tissue reweighting.}
Variants: asymmetric $(+3,-\infty)$, symmetric $(+3,+3)$, none $(0,0)$. Symmetric has lower RI
than asymmetric in all 16 step-matched comparisons. At the asymmetric arm's seed-0 selected
checkpoint, asymmetric minus none RI and HEST are in Table~\ref{tab:ablations}(a). THUNDER $k$-NN, linear probing, and few-shot are lower with
asymmetric reweighting for every backbone except Phikon-v2 $k$-NN ($+0.1$). Only Phikon-v2
clears the RI tolerance $\epsilon$, and on HEST only Phikon-v2, Midnight-12k and H-optimus-0
exceed the tuned across-seed spread. The current recipe does not claim reweighting as a
general positive mechanism.

\paragraph{Batch composition.}
Phikon-v2, one seed, a 750-step diagnostic sweep at a fixed 3,600-image budget with $CT$ held
constant. Table~\ref{tab:ablations}(b) reports RI at step 750 as the number of conditions per
batch $C$ decreases, and, for the two $C$ with a second $T$ run, the change in RI from
increasing the number of locations $T$ at fixed $C$, at three steps. Because $C$ and $T$ covary
at fixed $CT$, the first block is an allocation comparison rather than an isolated effect of
$C$; the recipe uses $C=2$.

\paragraph{Adapter rank.}
Mean RI over checkpoints at the stated steps, at fixed $\alpha/r=2$
(Table~\ref{tab:ablations}(c)). No monotone trend is present on Phikon-v2; increasing rank on Virchow2 leads to a monotone decrease, so the recipe uses rank 32. This sweep and the batch-composition sweep both
average over steps beyond the $100$--$200$ range where the recipe's checkpoints are selected
(\S\ref{sec:selection}) and are one-seed diagnostics.

\begin{table}[htbp]
\caption{Ablations, one seed each: (a) pseudo-tissue reweighting, (b) batch composition,
(c) adapter rank. Differences are read against the paper's
RI tolerance $\epsilon=0.007$ and, on HEST, against the tuned across-seed $2$\,SD of the same
backbone in the rightmost column of block (a); $^\dagger$ marks $|\Delta\mathrm{RI}|<\epsilon$,
and \textbf{bold} marks the recipe's setting.}
\label{tab:ablations}
\centering
\footnotesize
\setlength{\tabcolsep}{6pt}
\input{tables/ablations}
\end{table}

\clearpage
\section{Implementation details}
\label{app:implementation}

\paragraph{Adapters and heads.}
LoRA rank 32, $\alpha=64$, dropout 0, on every attention projection and MLP layer. Projection
heads are two-layer MLPs. GeM exponent initialized to 3, constrained to $[1,8]$. $\tau=0.07$.
Trainable parameters are $2.7$--$4.4\%$ of the backbone parameters for the LoRA adapters
alone, i.e.\ the parameters actually merged into the released backbone (Phikon-v2 $4.41$,
Midnight-12k $3.01$, Virchow2 and Virchow $3.20$, H-optimus-0, and OpenMidnight 1.1 $2.69$,
UNI2-h $2.68$), and $3.0$--$5.4\%$ for the adapters plus the two projection heads
(Phikon-v2 $5.40$, Midnight-12k $3.37$, Virchow2 $3.76$, H-optimus-0 $3.05$, UNI2-h $3.28$,
Virchow $3.76$, and OpenMidnight 1.1  $3.05$); the projection heads are discarded before merging.

\paragraph{Optimization and preprocessing.}
Native $224\times224$ tiles, no resizing, cropping, colour, or geometric augmentation; each
backbone's published channel normalization. AdamW, learning rate $10^{-4}$, weight decay $0.05$,
200 linear warmup steps, cosine decay over 500 steps, gradient clipping at $1.0$. bfloat16, gradient checkpointing,
activation offloading. Checkpoints every 50 steps. Three seeds per backbone controlling
initialization and sampling.

\paragraph{Compute.}
Two H100 80GB per job: one trains, one runs concurrent RI evaluation. 500-step runs take
$1.0$--$7.8$ hours per backbone. Evaluation dominates total compute.

%% file: tables/pathorob_submetrics.tex
\begin{tabular}{llcccccc}
\toprule
Backbone & Dataset & $n$ & Bal. acc. & ID purity & OOD purity & Pooled purity & RI \\
\midrule
Phikon-v2 & TCGA $2\times2$ & 3 & 0.921$\to$\loss{0.906 $\pm$ 0.003} & 0.871$\to$\loss{0.842 $\pm$ 0.006} & 0.887$\to$\loss{0.851 $\pm$ 0.007} & 0.874$\to$\loss{0.846 $\pm$ 0.006} & 0.617$\to$\gain{0.793 $\pm$ 0.006} \\
 & Camelyon & 3 & 0.955$\to$\gain{0.965 $\pm$ 0.001} & 0.942$\to$\gain{0.949 $\pm$ 0.001} & 0.994$\to$\loss{0.960 $\pm$ 0.000} & 0.942$\to$\gain{0.951 $\pm$ 0.001} & 0.023$\to$\gain{0.775 $\pm$ 0.040} \\
 & Tolkach-ESCA & 3 & 0.958$\to$0.961 $\pm$ 0.006 & 0.924$\to$0.918 $\pm$ 0.010 & 0.948$\to$\loss{0.928 $\pm$ 0.009} & 0.929$\to$0.924 $\pm$ 0.009 & 0.770$\to$\gain{0.939 $\pm$ 0.005} \\
 & Mean & 3 & 0.944$\to$0.944 $\pm$ 0.003 & 0.913$\to$\loss{0.903 $\pm$ 0.006} & 0.943$\to$\loss{0.913 $\pm$ 0.005} & 0.915$\to$\loss{0.907 $\pm$ 0.005} & 0.470$\to$\gain{0.836 $\pm$ 0.014} \\
\midrule
Midnight-12k & TCGA $2\times2$ & 3 & 0.934$\to$\gain{0.943 $\pm$ 0.005} & 0.916$\to$\loss{0.902 $\pm$ 0.013} & 0.911$\to$0.908 $\pm$ 0.011 & 0.914$\to$0.904 $\pm$ 0.012 & 0.858$\to$\gain{0.871 $\pm$ 0.010} \\
 & Camelyon & 3 & 0.976$\to$0.975 $\pm$ 0.003 & 0.966$\to$0.966 $\pm$ 0.003 & 0.964$\to$\gain{0.970 $\pm$ 0.003} & 0.966$\to$0.967 $\pm$ 0.003 & 0.478$\to$\gain{0.884 $\pm$ 0.019} \\
 & Tolkach-ESCA & 3 & 0.976$\to$0.977 $\pm$ 0.001 & 0.958$\to$\loss{0.953 $\pm$ 0.003} & 0.961$\to$\gain{0.965 $\pm$ 0.002} & 0.959$\to$0.960 $\pm$ 0.002 & 0.941$\to$\gain{0.969 $\pm$ 0.002} \\
 & Mean & 3 & 0.962$\to$\gain{0.965 $\pm$ 0.002} & 0.947$\to$0.940 $\pm$ 0.006 & 0.945$\to$0.948 $\pm$ 0.005 & 0.946$\to$0.944 $\pm$ 0.006 & 0.759$\to$\gain{0.908 $\pm$ 0.005} \\
\midrule
Virchow2 & TCGA $2\times2$ & 3 & 0.928$\to$0.924 $\pm$ 0.004 & 0.886$\to$\loss{0.876 $\pm$ 0.009} & 0.905$\to$\loss{0.887 $\pm$ 0.010} & 0.893$\to$\loss{0.880 $\pm$ 0.009} & 0.822$\to$\gain{0.841 $\pm$ 0.007} \\
 & Camelyon & 3 & 0.988$\to$\loss{0.982 $\pm$ 0.002} & 0.981$\to$\loss{0.974 $\pm$ 0.002} & 0.986$\to$\loss{0.976 $\pm$ 0.003} & 0.981$\to$\loss{0.974 $\pm$ 0.002} & 0.806$\to$\gain{0.916 $\pm$ 0.019} \\
 & Tolkach-ESCA & 3 & 0.977$\to$\gain{0.980 $\pm$ 0.001} & 0.949$\to$0.953 $\pm$ 0.005 & 0.964$\to$0.966 $\pm$ 0.002 & 0.957$\to$\gain{0.961 $\pm$ 0.003} & 0.955$\to$\gain{0.969 $\pm$ 0.001} \\
 & Mean & 3 & 0.964$\to$0.962 $\pm$ 0.002 & 0.939$\to$0.934 $\pm$ 0.005 & 0.952$\to$\loss{0.943 $\pm$ 0.004} & 0.944$\to$0.939 $\pm$ 0.005 & 0.861$\to$\gain{0.909 $\pm$ 0.007} \\
\midrule
H-optimus-0 & TCGA $2\times2$ & 3 & 0.939$\to$\loss{0.926 $\pm$ 0.003} & 0.895$\to$\loss{0.870 $\pm$ 0.008} & 0.917$\to$\loss{0.877 $\pm$ 0.008} & 0.902$\to$\loss{0.873 $\pm$ 0.008} & 0.802$\to$\gain{0.836 $\pm$ 0.007} \\
 & Camelyon & 3 & 0.982$\to$0.983 $\pm$ 0.001 & 0.973$\to$0.972 $\pm$ 0.002 & 0.969$\to$\gain{0.973 $\pm$ 0.002} & 0.973$\to$0.973 $\pm$ 0.002 & 0.678$\to$\gain{0.919 $\pm$ 0.004} \\
 & Tolkach-ESCA & 3 & 0.972$\to$0.973 $\pm$ 0.001 & 0.942$\to$0.941 $\pm$ 0.003 & 0.963$\to$\loss{0.948 $\pm$ 0.003} & 0.951$\to$\loss{0.945 $\pm$ 0.003} & 0.918$\to$\gain{0.962 $\pm$ 0.001} \\
 & Mean & 3 & 0.964$\to$\loss{0.961 $\pm$ 0.002} & 0.937$\to$\loss{0.928 $\pm$ 0.004} & 0.950$\to$\loss{0.933 $\pm$ 0.004} & 0.942$\to$\loss{0.930 $\pm$ 0.004} & 0.800$\to$\gain{0.906 $\pm$ 0.004} \\
\midrule
UNI2-h & TCGA $2\times2$ & 3 & 0.942$\to$\loss{0.932 $\pm$ 0.009} & 0.909$\to$\loss{0.889 $\pm$ 0.018} & 0.935$\to$\loss{0.900 $\pm$ 0.021} & 0.916$\to$\loss{0.894 $\pm$ 0.019} & 0.803$\to$\gain{0.858 $\pm$ 0.007} \\
 & Camelyon & 3 & 0.985$\to$0.983 $\pm$ 0.005 & 0.980$\to$0.975 $\pm$ 0.007 & 0.995$\to$\loss{0.981 $\pm$ 0.006} & 0.980$\to$0.976 $\pm$ 0.006 & 0.544$\to$\gain{0.901 $\pm$ 0.021} \\
 & Tolkach-ESCA & 3 & 0.976$\to$\loss{0.974 $\pm$ 0.001} & 0.951$\to$0.949 $\pm$ 0.005 & 0.974$\to$\loss{0.962 $\pm$ 0.009} & 0.959$\to$0.956 $\pm$ 0.006 & 0.923$\to$\gain{0.963 $\pm$ 0.006} \\
 & Mean & 3 & 0.968$\to$0.963 $\pm$ 0.005 & 0.947$\to$0.938 $\pm$ 0.010 & 0.968$\to$\loss{0.948 $\pm$ 0.012} & 0.952$\to$0.942 $\pm$ 0.010 & 0.757$\to$\gain{0.907 $\pm$ 0.007} \\
\midrule
Virchow & TCGA $2\times2$ & 3 & 0.909$\to$\loss{0.901 $\pm$ 0.006} & 0.866$\to$\loss{0.833 $\pm$ 0.014} & 0.871$\to$\loss{0.842 $\pm$ 0.015} & 0.867$\to$\loss{0.837 $\pm$ 0.014} & 0.761$\to$\gain{0.807 $\pm$ 0.009} \\
 & Camelyon & 3 & 0.980$\to$0.979 $\pm$ 0.002 & 0.975$\to$\loss{0.968 $\pm$ 0.003} & 0.932$\to$\gain{0.973 $\pm$ 0.003} & 0.972$\to$0.969 $\pm$ 0.003 & 0.751$\to$\gain{0.906 $\pm$ 0.002} \\
 & Tolkach-ESCA & 3 & 0.969$\to$\loss{0.965 $\pm$ 0.003} & 0.945$\to$\loss{0.927 $\pm$ 0.008} & 0.963$\to$\loss{0.935 $\pm$ 0.007} & 0.953$\to$\loss{0.932 $\pm$ 0.007} & 0.932$\to$\gain{0.958 $\pm$ 0.003} \\
 & Mean & 3 & 0.953$\to$\loss{0.948 $\pm$ 0.004} & 0.929$\to$\loss{0.909 $\pm$ 0.008} & 0.922$\to$0.917 $\pm$ 0.008 & 0.931$\to$\loss{0.913 $\pm$ 0.008} & 0.815$\to$\gain{0.890 $\pm$ 0.004} \\
\midrule
OpenMidnight 1.1 & TCGA $2\times2$ & 3 & 0.930$\to$\loss{0.921 $\pm$ 0.006} & 0.895$\to$\loss{0.866 $\pm$ 0.016} & 0.908$\to$\loss{0.876 $\pm$ 0.016} & 0.898$\to$\loss{0.870 $\pm$ 0.016} & 0.771$\to$\gain{0.834 $\pm$ 0.012} \\
 & Camelyon & 3 & 0.965$\to$\gain{0.974 $\pm$ 0.003} & 0.956$\to$\gain{0.962 $\pm$ 0.005} & 0.947$\to$\gain{0.965 $\pm$ 0.008} & 0.955$\to$\gain{0.963 $\pm$ 0.005} & 0.111$\to$\gain{0.862 $\pm$ 0.016} \\
 & Tolkach-ESCA & 3 & 0.975$\to$0.975 $\pm$ 0.002 & 0.954$\to$\loss{0.948 $\pm$ 0.005} & 0.966$\to$0.958 $\pm$ 0.008 & 0.958$\to$0.954 $\pm$ 0.007 & 0.916$\to$\gain{0.968 $\pm$ 0.000} \\
 & Mean & 3 & 0.957$\to$0.956 $\pm$ 0.003 & 0.935$\to$\loss{0.925 $\pm$ 0.008} & 0.940$\to$0.933 $\pm$ 0.011 & 0.937$\to$0.929 $\pm$ 0.009 & 0.600$\to$\gain{0.888 $\pm$ 0.004} \\
\bottomrule
\end{tabular}

%% file: tables/perseed_pathorob.tex
\begin{tabular}{lcccc}
\toprule
Backbone & Base & Seed a & Seed b & Seed c \\
\midrule
Phikon-v2 & 0.470 & 0.842 & 0.836 & 0.828 \\
Midnight-12k & 0.759 & 0.911 & 0.906 & 0.908 \\
Virchow2 & 0.861 & 0.905 & 0.913 & 0.909 \\
H-optimus-0 & 0.800 & 0.905 & 0.904 & 0.908 \\
UNI2-h & 0.757 & 0.905 & 0.911 & 0.906 \\
Virchow & 0.815 & 0.892 & 0.891 & 0.888 \\
OpenMidnight 1.1 & 0.600 & 0.889 & 0.889 & 0.885 \\
\bottomrule
\end{tabular}

%% file: tables/hest_per_task.tex
\begin{tabular}{lccccccccccc}
\toprule
 & \multicolumn{9}{c}{per-task mean Pearson $r$\,$\uparrow$ $\pm$ 2\,SD (rank\,$\downarrow$), base $\to$ fine-tuned} & & \\
\cmidrule(lr){2-10}
Backbone & IDC & PRAD & PAAD & SKCM & COAD & READ & CCRCC & LUNG & LYMPH\_IDC & Mean $r$\,$\uparrow$ & $\Sigma$rank\,(position)\,$\downarrow$ \\
\midrule
Phikon-v2 & \shortstack{0.541$\to$\gain{0.560$\pm$0.002} \\ {\scriptsize (18$\to$\gain{14})}} & \shortstack{0.354$\to$\loss{0.340$\pm$0.003} \\ {\scriptsize (12$\to$\loss{18})}} & \shortstack{0.446$\to$\gain{0.480$\pm$0.007} \\ {\scriptsize (17$\to$\gain{9})}} & \shortstack{0.555$\to$\gain{0.587$\pm$0.013} \\ {\scriptsize (17$\to$\gain{12})}} & \shortstack{0.250$\to$\gain{0.297$\pm$0.017} \\ {\scriptsize (18$\to$\gain{8})}} & \shortstack{0.175$\to$0.166$\pm$0.010 \\ {\scriptsize (13$\to$13)}} & \shortstack{0.266$\to$0.261$\pm$0.017 \\ {\scriptsize (4$\to$\loss{7})}} & \shortstack{0.542$\to$\gain{0.566$\pm$0.005} \\ {\scriptsize (15$\to$\gain{6})}} & \shortstack{0.244$\to$\gain{0.259$\pm$0.002} \\ {\scriptsize (17$\to$\gain{10})}} & 0.3747$\to$\gain{0.3906$\pm$0.0006} & 131~(15)$\to$\gain{97~(10)} \\
\addlinespace[2pt]
Midnight-12k & \shortstack{0.582$\to$\gain{0.588$\pm$0.004} \\ {\scriptsize (10$\to$\gain{6})}} & \shortstack{0.337$\to$\gain{0.368$\pm$0.010} \\ {\scriptsize (20$\to$\gain{7})}} & \shortstack{0.490$\to$\gain{0.517$\pm$0.010} \\ {\scriptsize (7$\to$\gain{1})}} & \shortstack{0.636$\to$\loss{0.633$\pm$0.003} \\ {\scriptsize (6$\to$\loss{7})}} & \shortstack{0.291$\to$\gain{0.325$\pm$0.008} \\ {\scriptsize (8$\to$\gain{2})}} & \shortstack{0.185$\to$0.183$\pm$0.008 \\ {\scriptsize (9$\to$\loss{10})}} & \shortstack{0.213$\to$\gain{0.262$\pm$0.014} \\ {\scriptsize (24$\to$\gain{7})}} & \shortstack{0.558$\to$0.565$\pm$0.008 \\ {\scriptsize (8$\to$\gain{6})}} & \shortstack{0.264$\to$\gain{0.269$\pm$0.002} \\ {\scriptsize (5$\to$5)}} & 0.3952$\to$\gain{0.4122$\pm$0.0022} & 97~(8)$\to$\gain{51~(5)} \\
\addlinespace[2pt]
Virchow2 & \shortstack{0.597$\to$\loss{0.591$\pm$0.001} \\ {\scriptsize (3$\to$3)}} & \shortstack{0.353$\to$\gain{0.381$\pm$0.024} \\ {\scriptsize (13$\to$\gain{4})}} & \shortstack{0.478$\to$\gain{0.503$\pm$0.015} \\ {\scriptsize (9$\to$\gain{2})}} & \shortstack{0.639$\to$0.636$\pm$0.009 \\ {\scriptsize (5$\to$\loss{6})}} & \shortstack{0.258$\to$\gain{0.283$\pm$0.014} \\ {\scriptsize (16$\to$\gain{9})}} & \shortstack{0.207$\to$0.203$\pm$0.019 \\ {\scriptsize (4$\to$4)}} & \shortstack{0.272$\to$\loss{0.262$\pm$0.006} \\ {\scriptsize (1$\to$\loss{6})}} & \shortstack{0.568$\to$0.563$\pm$0.013 \\ {\scriptsize (4$\to$\loss{5})}} & \shortstack{0.257$\to$0.259$\pm$0.004 \\ {\scriptsize (10$\to$\gain{9})}} & 0.4032$\to$\gain{0.4089$\pm$0.0034} & 65~(6)$\to$\gain{48~(5)} \\
\addlinespace[2pt]
H-optimus-0 & \shortstack{0.598$\to$\gain{0.602$\pm$0.004} \\ {\scriptsize (2$\to$2)}} & \shortstack{0.385$\to$\gain{0.389$\pm$0.003} \\ {\scriptsize (2$\to$2)}} & \shortstack{0.491$\to$\gain{0.507$\pm$0.004} \\ {\scriptsize (6$\to$\gain{2})}} & \shortstack{0.645$\to$\gain{0.658$\pm$0.005} \\ {\scriptsize (4$\to$4)}} & \shortstack{0.309$\to$\loss{0.287$\pm$0.004} \\ {\scriptsize (3$\to$\loss{8})}} & \shortstack{0.222$\to$\gain{0.235$\pm$0.002} \\ {\scriptsize (3$\to$\gain{2})}} & \shortstack{0.268$\to$\gain{0.280$\pm$0.006} \\ {\scriptsize (2$\to$\gain{1})}} & \shortstack{0.559$\to$\gain{0.572$\pm$0.003} \\ {\scriptsize (6$\to$\gain{4})}} & \shortstack{0.259$\to$\gain{0.272$\pm$0.002} \\ {\scriptsize (9$\to$\gain{4})}} & 0.4150$\to$\gain{0.4226$\pm$0.0018} & 37~(3)$\to$\gain{29~(2)} \\
\addlinespace[2pt]
UNI2-h & \shortstack{0.590$\to$\gain{0.598$\pm$0.004} \\ {\scriptsize (4$\to$\gain{2})}} & \shortstack{0.357$\to$\gain{0.382$\pm$0.012} \\ {\scriptsize (9$\to$\gain{3})}} & \shortstack{0.500$\to$0.512$\pm$0.012 \\ {\scriptsize (2$\to$2)}} & \shortstack{0.659$\to$0.665$\pm$0.010 \\ {\scriptsize (2$\to$2)}} & \shortstack{0.301$\to$\gain{0.329$\pm$0.004} \\ {\scriptsize (6$\to$\gain{1})}} & \shortstack{0.223$\to$0.222$\pm$0.015 \\ {\scriptsize (2$\to$2)}} & \shortstack{0.264$\to$0.271$\pm$0.032 \\ {\scriptsize (6$\to$\gain{2})}} & \shortstack{0.558$\to$0.564$\pm$0.013 \\ {\scriptsize (8$\to$\gain{6})}} & \shortstack{0.272$\to$0.269$\pm$0.011 \\ {\scriptsize (3$\to$\loss{4})}} & 0.4138$\to$\gain{0.4238$\pm$0.0058} & 42~(4)$\to$\gain{24~(1)} \\
\addlinespace[2pt]
Virchow & \shortstack{0.585$\to$0.585$\pm$0.002 \\ {\scriptsize (9$\to$9)}} & \shortstack{0.338$\to$\gain{0.382$\pm$0.006} \\ {\scriptsize (19$\to$\gain{3})}} & \shortstack{0.516$\to$\loss{0.507$\pm$0.003} \\ {\scriptsize (1$\to$1)}} & \shortstack{0.625$\to$\gain{0.640$\pm$0.012} \\ {\scriptsize (8$\to$\gain{6})}} & \shortstack{0.308$\to$0.301$\pm$0.015 \\ {\scriptsize (4$\to$\loss{6})}} & \shortstack{0.198$\to$0.189$\pm$0.009 \\ {\scriptsize (5$\to$\loss{7})}} & \shortstack{0.259$\to$0.255$\pm$0.008 \\ {\scriptsize (9$\to$9)}} & \shortstack{0.566$\to$\loss{0.555$\pm$0.005} \\ {\scriptsize (5$\to$\loss{8})}} & \shortstack{0.261$\to$0.261$\pm$0.003 \\ {\scriptsize (7$\to$7)}} & 0.4061$\to$0.4083$\pm$0.0036 & 67~(5)$\to$\gain{56~(5)} \\
\addlinespace[2pt]
OpenMidnight 1.1 & \shortstack{0.577$\to$0.579$\pm$0.010 \\ {\scriptsize (11$\to$11)}} & \shortstack{0.349$\to$\gain{0.381$\pm$0.007} \\ {\scriptsize (14$\to$\gain{3})}} & \shortstack{0.475$\to$\gain{0.500$\pm$0.005} \\ {\scriptsize (10$\to$\gain{2})}} & \shortstack{0.606$\to$\loss{0.592$\pm$0.013} \\ {\scriptsize (9$\to$\loss{11})}} & \shortstack{0.223$\to$\gain{0.280$\pm$0.023} \\ {\scriptsize (25$\to$\gain{12})}} & \shortstack{0.155$\to$\gain{0.168$\pm$0.005} \\ {\scriptsize (19$\to$\gain{14})}} & \shortstack{0.257$\to$0.254$\pm$0.004 \\ {\scriptsize (10$\to$10)}} & \shortstack{0.551$\to$\loss{0.538$\pm$0.006} \\ {\scriptsize (11$\to$\loss{18})}} & \shortstack{0.257$\to$\gain{0.263$\pm$0.001} \\ {\scriptsize (10$\to$\gain{7})}} & 0.3834$\to$\gain{0.3950$\pm$0.0026} & 119~(13)$\to$\gain{88~(9)} \\
\bottomrule
\end{tabular}

%% file: tables/perseed_hest.tex
\begin{tabular}{lcccc}
\toprule
Backbone & Base & Seed a & Seed b & Seed c \\
\midrule
Phikon-v2 & 0.375 & 0.390 & 0.391 & 0.391 \\
Midnight-12k & 0.395 & 0.413 & 0.412 & 0.411 \\
Virchow2 & 0.403 & 0.409 & 0.410 & 0.407 \\
H-optimus-0 & 0.415 & 0.421 & 0.423 & 0.423 \\
UNI2-h & 0.414 & 0.427 & 0.421 & 0.424 \\
Virchow & 0.406 & 0.407 & 0.407 & 0.410 \\
OpenMidnight 1.1 & 0.383 & 0.397 & 0.394 & 0.394 \\
\bottomrule
\end{tabular}

%% file: tables/cptac_per_task.tex
\begin{tabular}{lccccccc}
\toprule
Task & Phikon-v2 & Midnight-12k & Virchow2 & H-optimus-0 & UNI2-h & Virchow & OpenMidnight 1.1 \\
\midrule
BRCA / Immune class & 0.718$\to$\gain{0.748} & 0.708$\to$\gain{0.732} & 0.728$\to$\gain{0.737} & 0.730$\to$\gain{0.737} & 0.708$\to$\gain{0.744} & 0.714$\to$\gain{0.726} & 0.703$\to$\gain{0.743} \\
BRCA / PIK3CA mutation & 0.546$\to$\gain{0.568} & 0.531$\to$\gain{0.578} & 0.551$\to$0.563 & 0.559$\to$\loss{0.542} & 0.559$\to$\gain{0.586} & 0.576$\to$\loss{0.505} & 0.542$\to$0.544 \\
BRCA / TP53 mutation & 0.832$\to$\loss{0.802} & 0.811$\to$\loss{0.773} & 0.762$\to$0.762 & 0.775$\to$\loss{0.758} & 0.786$\to$0.798 & 0.781$\to$\loss{0.766} & 0.780$\to$\loss{0.770} \\
\midrule
CCRCC / BAP1 mutation & 0.638$\to$\gain{0.785} & 0.669$\to$\gain{0.738} & 0.650$\to$\gain{0.706} & 0.681$\to$\gain{0.722} & 0.742$\to$\gain{0.763} & 0.598$\to$\gain{0.695} & 0.636$\to$\gain{0.730} \\
CCRCC / Immune class & 0.483$\to$\gain{0.503} & 0.538$\to$\gain{0.554} & 0.551$\to$0.541 & 0.550$\to$0.552 & 0.547$\to$0.534 & 0.537$\to$\gain{0.555} & 0.583$\to$\loss{0.543} \\
CCRCC / PBRM1 mutation & 0.456$\to$\gain{0.491} & 0.530$\to$\gain{0.538} & 0.477$\to$\gain{0.490} & 0.437$\to$\gain{0.477} & 0.451$\to$\gain{0.517} & 0.476$\to$\gain{0.490} & 0.520$\to$\loss{0.480} \\
CCRCC / VHL mutation & 0.561$\to$\gain{0.614} & 0.540$\to$\gain{0.560} & 0.618$\to$\loss{0.588} & 0.606$\to$\loss{0.581} & 0.540$\to$\gain{0.544} & 0.592$\to$\gain{0.616} & 0.556$\to$0.576 \\
\midrule
COAD / ACVR2A mutation & 0.723$\to$\gain{0.766} & 0.678$\to$\gain{0.771} & 0.750$\to$\gain{0.795} & 0.775$\to$\gain{0.810} & 0.769$\to$0.772 & 0.672$\to$\gain{0.768} & 0.716$\to$\gain{0.761} \\
COAD / APC mutation & 0.703$\to$\gain{0.716} & 0.741$\to$\loss{0.686} & 0.666$\to$\gain{0.686} & 0.691$\to$\gain{0.712} & 0.729$\to$0.729 & 0.711$\to$\gain{0.722} & 0.690$\to$\gain{0.718} \\
COAD / ARID1A mutation & 0.799$\to$\gain{0.857} & 0.801$\to$\gain{0.845} & 0.808$\to$\gain{0.821} & 0.814$\to$\gain{0.837} & 0.767$\to$\gain{0.831} & 0.712$\to$\gain{0.849} & 0.828$\to$0.835 \\
COAD / Immune class & 0.605$\to$\gain{0.640} & 0.580$\to$\gain{0.671} & 0.650$\to$\gain{0.677} & 0.609$\to$\gain{0.647} & 0.602$\to$\gain{0.652} & 0.598$\to$\gain{0.662} & 0.586$\to$\gain{0.653} \\
COAD / KRAS mutation & 0.643$\to$0.645 & 0.618$\to$0.617 & 0.608$\to$\gain{0.639} & 0.599$\to$\gain{0.627} & 0.657$\to$\loss{0.645} & 0.634$\to$0.633 & 0.626$\to$\loss{0.616} \\
COAD / MSI H & 0.819$\to$\gain{0.885} & 0.791$\to$\gain{0.860} & 0.868$\to$\gain{0.910} & 0.905$\to$\gain{0.925} & 0.881$\to$\gain{0.898} & 0.797$\to$\gain{0.894} & 0.764$\to$\gain{0.872} \\
COAD / PIK3CA mutation & 0.655$\to$\gain{0.710} & 0.710$\to$\gain{0.752} & 0.704$\to$\gain{0.742} & 0.703$\to$\gain{0.755} & 0.677$\to$\gain{0.700} & 0.617$\to$\gain{0.719} & 0.691$\to$\gain{0.735} \\
COAD / SETD1B mutation & 0.712$\to$\gain{0.776} & 0.765$\to$0.765 & 0.822$\to$\gain{0.851} & 0.825$\to$\gain{0.860} & 0.794$\to$\gain{0.816} & 0.715$\to$\gain{0.841} & 0.714$\to$\gain{0.760} \\
COAD / TP53 mutation & 0.714$\to$\loss{0.684} & 0.599$\to$\gain{0.670} & 0.727$\to$\loss{0.686} & 0.693$\to$0.684 & 0.745$\to$\loss{0.700} & 0.665$\to$\gain{0.720} & 0.686$\to$\loss{0.640} \\
\midrule
GBM / EGFR mutation & 0.585$\to$\gain{0.652} & 0.578$\to$\gain{0.640} & 0.536$\to$\gain{0.593} & 0.539$\to$\gain{0.589} & 0.587$\to$\gain{0.640} & 0.501$\to$\gain{0.575} & 0.524$\to$\gain{0.616} \\
GBM / Immune class & 0.584$\to$\gain{0.693} & 0.716$\to$\gain{0.732} & 0.683$\to$\gain{0.696} & 0.703$\to$\loss{0.693} & 0.742$\to$0.740 & 0.712$\to$\gain{0.738} & 0.671$\to$\gain{0.688} \\
GBM / TP53 mutation & 0.718$\to$\gain{0.777} & 0.756$\to$\gain{0.773} & 0.790$\to$\gain{0.841} & 0.825$\to$\gain{0.862} & 0.800$\to$0.825 & 0.818$\to$\loss{0.801} & 0.799$\to$0.794 \\
\midrule
HNSC / CASP8 mutation & 0.449$\to$\gain{0.588} & 0.653$\to$0.654 & 0.567$\to$\loss{0.540} & 0.463$\to$\gain{0.492} & 0.438$\to$\gain{0.579} & 0.446$\to$0.467 & 0.465$\to$\gain{0.522} \\
HNSC / Histologic Grade & 0.751$\to$0.751 & 0.797$\to$\loss{0.764} & 0.757$\to$\loss{0.743} & 0.734$\to$\gain{0.747} & 0.782$\to$\loss{0.760} & 0.755$\to$\loss{0.713} & 0.797$\to$\loss{0.778} \\
HNSC / Immune class & 0.608$\to$\gain{0.680} & 0.670$\to$\gain{0.699} & 0.622$\to$\gain{0.700} & 0.653$\to$\gain{0.690} & 0.669$\to$\gain{0.706} & 0.638$\to$\gain{0.681} & 0.660$\to$\gain{0.682} \\
\midrule
LSCC / ARID1A mutation & 0.395$\to$\gain{0.448} & 0.377$\to$\gain{0.433} & 0.370$\to$\loss{0.345} & 0.374$\to$\gain{0.385} & 0.397$\to$\gain{0.433} & 0.430$\to$\loss{0.415} & 0.398$\to$\gain{0.411} \\
LSCC / Histologic Grade & 0.708$\to$\loss{0.623} & 0.630$\to$\loss{0.608} & 0.648$\to$0.629 & 0.692$\to$\loss{0.637} & 0.686$\to$0.670 & 0.682$\to$\loss{0.606} & 0.695$\to$\loss{0.650} \\
LSCC / Immune class & 0.682$\to$\gain{0.762} & 0.746$\to$\gain{0.777} & 0.755$\to$0.755 & 0.727$\to$\gain{0.764} & 0.686$\to$\gain{0.758} & 0.737$\to$\gain{0.765} & 0.740$\to$\gain{0.764} \\
LSCC / KEAP1 mutation & 0.665$\to$\gain{0.710} & 0.680$\to$\gain{0.709} & 0.712$\to$0.706 & 0.660$\to$\gain{0.683} & 0.718$\to$\loss{0.712} & 0.640$\to$\gain{0.664} & 0.653$\to$0.662 \\
\midrule
LUAD / EGFR mutation & 0.740$\to$\gain{0.747} & 0.764$\to$\loss{0.745} & 0.821$\to$\loss{0.802} & 0.797$\to$\loss{0.782} & 0.789$\to$0.784 & 0.795$\to$\loss{0.767} & 0.783$\to$\loss{0.750} \\
LUAD / Immune class & 0.600$\to$\gain{0.716} & 0.666$\to$\gain{0.732} & 0.653$\to$\gain{0.668} & 0.616$\to$\gain{0.671} & 0.625$\to$\gain{0.686} & 0.663$\to$\gain{0.707} & 0.646$\to$\gain{0.702} \\
LUAD / KRAS mutation & 0.660$\to$\loss{0.643} & 0.630$\to$\gain{0.659} & 0.711$\to$0.711 & 0.690$\to$\gain{0.714} & 0.678$\to$\gain{0.708} & 0.676$\to$\loss{0.654} & 0.637$\to$0.630 \\
LUAD / STK11 mutation & 0.828$\to$\gain{0.842} & 0.834$\to$\gain{0.856} & 0.818$\to$\gain{0.845} & 0.867$\to$\gain{0.871} & 0.868$\to$\gain{0.884} & 0.843$\to$\gain{0.851} & 0.857$\to$0.858 \\
LUAD / TP53 mutation & 0.703$\to$0.698 & 0.707$\to$0.701 & 0.734$\to$\loss{0.702} & 0.723$\to$0.720 & 0.742$\to$0.753 & 0.738$\to$\loss{0.710} & 0.717$\to$0.703 \\
\midrule
LUNG / subtype & 0.987$\to$0.987 & 0.992$\to$\gain{0.994} & 0.992$\to$\loss{0.990} & 0.995$\to$0.995 & 0.995$\to$0.994 & 0.987$\to$0.986 & 0.990$\to$0.989 \\
\midrule
OV / Immune class & 0.558$\to$\gain{0.587} & 0.483$\to$\gain{0.542} & 0.516$\to$0.520 & 0.539$\to$\loss{0.534} & 0.535$\to$\gain{0.551} & 0.513$\to$\gain{0.522} & 0.561$\to$\loss{0.526} \\
\midrule
PDA / Immune class & 0.509$\to$\gain{0.568} & 0.601$\to$\gain{0.614} & 0.586$\to$\gain{0.589} & 0.566$\to$\gain{0.569} & 0.543$\to$\gain{0.578} & 0.586$\to$\gain{0.597} & 0.556$\to$\gain{0.567} \\
PDA / SMAD4 mutation & 0.379$\to$\gain{0.480} & 0.467$\to$0.472 & 0.475$\to$\gain{0.538} & 0.467$\to$\gain{0.521} & 0.396$\to$\gain{0.477} & 0.542$\to$0.540 & 0.421$\to$\gain{0.529} \\
\midrule
UCEC / CTNNB1 mutation & 0.779$\to$\loss{0.762} & 0.772$\to$0.762 & 0.763$\to$0.742 & 0.783$\to$\loss{0.769} & 0.739$\to$\gain{0.766} & 0.723$\to$\gain{0.764} & 0.724$\to$\gain{0.764} \\
UCEC / Immune class & 0.502$\to$\gain{0.566} & 0.524$\to$\gain{0.638} & 0.654$\to$0.651 & 0.594$\to$\gain{0.633} & 0.595$\to$\gain{0.634} & 0.649$\to$\gain{0.716} & 0.577$\to$\gain{0.625} \\
UCEC / PTEN mutation & 0.668$\to$\loss{0.636} & 0.592$\to$0.598 & 0.694$\to$\loss{0.641} & 0.606$\to$\gain{0.653} & 0.689$\to$0.657 & 0.641$\to$\loss{0.587} & 0.569$\to$0.567 \\
\midrule
Mean (38) & 0.649$\to$\gain{0.687} & 0.664$\to$\gain{0.690} & 0.679$\to$\gain{0.688} & 0.673$\to$\gain{0.689} & 0.675$\to$\gain{0.698} & 0.661$\to$\gain{0.684} & 0.660$\to$\gain{0.678} \\
\bottomrule
\end{tabular}

%% file: tables/perseed_cptac.tex
\begin{tabular}{lcccc}
\toprule
Backbone & Base & Seed a & Seed b & Seed c \\
\midrule
Phikon-v2 & 0.649 & 0.686 & 0.688 & 0.686 \\
Midnight-12k & 0.664 & 0.689 & 0.689 & 0.691 \\
Virchow2 & 0.679 & 0.687 & 0.688 & 0.689 \\
H-optimus-0 & 0.673 & 0.689 & 0.689 & 0.690 \\
UNI2-h & 0.675 & 0.700 & 0.694 & 0.699 \\
Virchow & 0.661 & 0.683 & 0.685 & 0.684 \\
OpenMidnight 1.1 & 0.660 & 0.676 & 0.679 & 0.678 \\
\bottomrule
\end{tabular}

%% file: tables/thunder_ranks_full.tex
\begin{tabular}{lccccccc}
\toprule
Backbone & $k$-NN\,$\uparrow$ & Lin.\ probe\,$\uparrow$ & Few-shot\,$\uparrow$ & Seg.\,$\uparrow$ & ECE\,$\downarrow$ & Adv.\ drop\,$\downarrow$ & $\Sigma$rank\,(position)\,$\downarrow$ \\
\midrule
Phikon-v2 & 73.9~(20)$\to$77.4$\pm$0.5~\gain{(18)} & 79.7~(19)$\to$81.4$\pm$0.3~\gain{(16)} & 71.8~(17)$\to$72.5$\pm$0.0~\gain{(16)} & 67.3~(11)$\to$67.3$\pm$0.5~(11) & 3.8~(8)$\to$3.9$\pm$0.2~\loss{(9)} & 43.6~(13)$\to$34.8$\pm$1.7~\gain{(5)} & 88~(18)$\to$\gain{75~(14)} \\
Midnight-12k & 80.0~(8)$\to$81.8$\pm$0.7~\gain{(5)} & 84.8~(4)$\to$85.6$\pm$0.2~\gain{(2)} & 71.5~(18)$\to$77.0$\pm$0.2~\gain{(6)} & 68.0~(9)$\to$68.7$\pm$1.0~\gain{(6)} & 2.9~(2)$\to$3.8$\pm$0.5~\loss{(7)} & 37.2~(6)$\to$29.1$\pm$5.4~\gain{(1)} & 47~(5)$\to$\gain{27~(2)} \\
Virchow2 & 82.9~(3)$\to$83.1$\pm$0.2~(3) & 84.7~(4)$\to$85.5$\pm$0.5~\gain{(2)} & 74.0~(12)$\to$78.1$\pm$0.6~\gain{(4)} & 69.0~(3)$\to$69.0$\pm$0.4~(3) & 4.0~(10)$\to$4.5$\pm$0.4~\loss{(14)} & 31.3~(1)$\to$25.4$\pm$0.6~(1) & 33~(3)$\to$\gain{27~(2)} \\
H-optimus-0 & 81.4~(6)$\to$81.5$\pm$0.5~\gain{(5)} & 83.8~(6)$\to$84.2$\pm$0.2~(6) & 76.1~(8)$\to$77.0$\pm$0.1~\gain{(6)} & 64.6~(14)$\to$65.0$\pm$0.9~(14) & 4.0~(10)$\to$3.5$\pm$0.2~\gain{(6)} & 44.1~(15)$\to$37.7$\pm$0.8~\gain{(7)} & 59~(8)$\to$\gain{44~(5)} \\
UNI2-h & 83.3~(2)$\to$83.0$\pm$0.8~(2) & 85.7~(1)$\to$85.8$\pm$0.6~(1) & 79.8~(1)$\to$78.8$\pm$1.0~\loss{(2)} & 69.2~(2)$\to$69.1$\pm$0.5~\loss{(3)} & 3.9~(9)$\to$4.5$\pm$0.2~\loss{(14)} & 31.6~(2)$\to$24.9$\pm$1.6~\gain{(1)} & 17~(1)$\to$\loss{23~(1)} \\
Virchow & 77.4~(18)$\to$78.9$\pm$0.6~\gain{(12)} & 82.8~(11)$\to$83.4$\pm$0.4~\gain{(9)} & 71.8~(17)$\to$73.7$\pm$0.4~\gain{(14)} & 69.2~(2)$\to$68.9$\pm$0.2~\loss{(4)} & 4.5~(14)$\to$4.0$\pm$0.5~\gain{(10)} & 38.4~(8)$\to$30.3$\pm$1.0~\gain{(1)} & 70~(11)$\to$\gain{50~(6)} \\
OpenMidnight 1.1 & 80.8~(7)$\to$81.1$\pm$1.1~(7) & 85.2~(3)$\to$84.8$\pm$0.2~\loss{(5)} & 74.0~(12)$\to$76.4$\pm$0.8~\gain{(7)} & 69.2~(2)$\to$69.0$\pm$0.4~\loss{(4)} & 5.8~(20)$\to$4.4$\pm$0.4~\gain{(14)} & 34.8~(5)$\to$26.3$\pm$1.5~\gain{(1)} & 49~(7)$\to$\gain{38~(3)} \\
\bottomrule
\end{tabular}

%% file: tables/thunder_per_ds_cls.tex
\begin{tabular}{lccccccc}
\toprule
Dataset & Phikon-v2 & Midnight-12k & Virchow2 & H-optimus-0 & UNI2-h & Virchow & OpenMidnight 1.1 \\
\midrule
\multicolumn{8}{l}{\emph{$k$-NN macro-F1\,$\uparrow$}} \\
\addlinespace[1pt]
bach & 53.1$\to$\gain{62.7$\pm$6.3} & 84.3$\to$84.0$\pm$3.7 & 81.6$\to$83.5$\pm$6.4 & 65.6$\to$\gain{70.2$\pm$1.3} & 85.3$\to$83.2$\pm$4.3 & 55.2$\to$\gain{64.7$\pm$2.5} & 74.5$\to$\gain{75.6$\pm$0.1} \\
bracs & 45.5$\to$\gain{51.4$\pm$2.0} & 50.2$\to$\gain{55.6$\pm$2.7} & 54.9$\to$57.7$\pm$3.0 & 52.2$\to$\gain{55.7$\pm$0.3} & 56.1$\to$57.2$\pm$2.4 & 51.5$\to$52.3$\pm$2.3 & 53.9$\to$55.7$\pm$6.5 \\
breakhis & 51.6$\to$\gain{63.5$\pm$1.0} & 58.1$\to$\gain{70.2$\pm$7.1} & 80.1$\to$78.1$\pm$3.3 & 81.9$\to$\loss{74.6$\pm$3.8} & 79.4$\to$77.9$\pm$2.1 & 56.2$\to$\gain{65.6$\pm$6.2} & 83.7$\to$81.8$\pm$4.7 \\
ccrcc & 77.2$\to$\gain{88.4$\pm$1.7} & 91.6$\to$\loss{90.5$\pm$0.8} & 91.1$\to$\gain{92.4$\pm$0.7} & 90.5$\to$\loss{89.8$\pm$0.3} & 90.9$\to$\gain{92.0$\pm$0.5} & 83.7$\to$\gain{90.4$\pm$0.4} & 91.8$\to$\gain{92.2$\pm$0.1} \\
crc & 92.1$\to$\gain{93.9$\pm$0.3} & 94.2$\to$\gain{94.7$\pm$0.3} & 94.3$\to$94.7$\pm$0.5 & 93.1$\to$\gain{94.8$\pm$0.9} & 95.0$\to$94.9$\pm$0.6 & 92.5$\to$\gain{95.3$\pm$0.6} & 94.6$\to$94.5$\pm$0.8 \\
esca & 75.9$\to$\gain{79.6$\pm$0.5} & 81.4$\to$\gain{83.3$\pm$0.6} & 86.6$\to$85.7$\pm$1.3 & 82.1$\to$82.7$\pm$0.9 & 83.2$\to$83.1$\pm$0.6 & 81.8$\to$\loss{80.0$\pm$0.3} & 79.6$\to$\gain{82.6$\pm$1.5} \\
mhist & 66.5$\to$\gain{71.1$\pm$2.1} & 69.6$\to$70.5$\pm$2.7 & 73.9$\to$\gain{77.9$\pm$2.0} & 73.2$\to$\gain{78.0$\pm$1.4} & 70.9$\to$73.9$\pm$3.8 & 68.2$\to$\gain{74.0$\pm$3.0} & 70.6$\to$\gain{76.5$\pm$2.2} \\
pcam & 82.2$\to$\gain{86.8$\pm$0.3} & 88.0$\to$\gain{89.8$\pm$0.6} & 89.7$\to$89.8$\pm$0.6 & 89.3$\to$89.1$\pm$0.4 & 93.4$\to$91.7$\pm$2.2 & 89.9$\to$89.4$\pm$0.5 & 86.0$\to$\gain{88.1$\pm$0.9} \\
spider\_br & 80.2$\to$\loss{77.3$\pm$0.3} & 77.1$\to$\gain{80.3$\pm$0.9} & 82.3$\to$\loss{81.3$\pm$0.6} & 81.8$\to$81.9$\pm$0.5 & 82.6$\to$82.1$\pm$2.2 & 79.3$\to$78.9$\pm$0.8 & 80.1$\to$\loss{77.7$\pm$2.1} \\
spider\_col & 86.1$\to$86.0$\pm$0.5 & 85.0$\to$\gain{86.9$\pm$0.8} & 88.0$\to$\gain{88.7$\pm$0.2} & 87.8$\to$87.6$\pm$0.5 & 87.1$\to$87.6$\pm$0.7 & 87.8$\to$\loss{87.1$\pm$0.6} & 85.9$\to$85.3$\pm$1.1 \\
spider\_sk & 83.3$\to$\loss{82.4$\pm$0.8} & 85.7$\to$84.8$\pm$0.9 & 89.1$\to$\loss{87.4$\pm$0.9} & 89.3$\to$\loss{87.8$\pm$0.7} & 90.5$\to$88.3$\pm$2.6 & 88.8$\to$\loss{85.0$\pm$1.2} & 85.5$\to$\loss{82.0$\pm$1.8} \\
spider\_th & 91.4$\to$92.4$\pm$1.1 & 92.7$\to$\gain{93.7$\pm$0.4} & 92.6$\to$\gain{94.3$\pm$0.8} & 93.8$\to$93.8$\pm$0.3 & 92.5$\to$\gain{93.6$\pm$0.5} & 92.3$\to$92.6$\pm$0.9 & 92.5$\to$93.0$\pm$0.8 \\
crc\_msi & 56.8$\to$\gain{62.4$\pm$0.5} & 61.9$\to$62.9$\pm$1.1 & 65.6$\to$65.3$\pm$0.3 & 62.4$\to$\gain{64.5$\pm$0.3} & 64.1$\to$\gain{65.6$\pm$0.4} & 61.8$\to$\gain{64.8$\pm$0.3} & 59.4$\to$\gain{64.5$\pm$0.4} \\
tils & 80.8$\to$\gain{87.7$\pm$0.4} & 87.7$\to$\gain{89.9$\pm$0.4} & 88.7$\to$88.5$\pm$0.9 & 89.1$\to$\loss{88.9$\pm$0.1} & 89.1$\to$89.0$\pm$0.1 & 88.7$\to$88.7$\pm$0.4 & 89.2$\to$\gain{90.4$\pm$0.3} \\
tcga\_unif & 69.1$\to$\loss{58.2$\pm$1.1} & 77.4$\to$\loss{74.9$\pm$2.1} & 70.9$\to$\loss{65.2$\pm$2.3} & 73.2$\to$\loss{66.7$\pm$1.6} & 75.1$\to$\loss{69.8$\pm$4.3} & 62.6$\to$\loss{57.1$\pm$1.9} & 72.7$\to$\loss{62.7$\pm$4.0} \\
wilds & 91.0$\to$\gain{95.0$\pm$0.3} & 95.0$\to$\gain{96.7$\pm$0.1} & 97.6$\to$\gain{98.0$\pm$0.1} & 97.9$\to$97.9$\pm$0.1 & 98.0$\to$\loss{97.6$\pm$0.2} & 97.5$\to$\loss{97.2$\pm$0.0} & 93.1$\to$\gain{96.0$\pm$0.2} \\
\midrule
\multicolumn{8}{l}{\emph{Linear-probe macro-F1\,$\uparrow$}} \\
\addlinespace[1pt]
bach & 64.7$\to$65.8$\pm$2.7 & 87.9$\to$\loss{82.3$\pm$0.0} & 76.5$\to$\gain{81.0$\pm$0.9} & 69.7$\to$70.4$\pm$2.5 & 84.7$\to$82.9$\pm$8.1 & 66.2$\to$67.5$\pm$2.4 & 85.0$\to$82.9$\pm$2.7 \\
bracs & 58.3$\to$58.3$\pm$0.6 & 63.9$\to$\loss{62.8$\pm$1.0} & 62.2$\to$60.8$\pm$2.1 & 61.1$\to$\gain{62.5$\pm$0.7} & 65.4$\to$64.7$\pm$2.9 & 59.9$\to$59.8$\pm$3.9 & 61.1$\to$\gain{63.8$\pm$0.9} \\
breakhis & 53.5$\to$\gain{64.1$\pm$4.3} & 56.7$\to$\gain{78.1$\pm$0.8} & 72.5$\to$\gain{76.0$\pm$2.2} & 66.0$\to$\gain{73.2$\pm$2.0} & 78.7$\to$\loss{75.6$\pm$2.3} & 62.7$\to$\gain{72.6$\pm$6.3} & 87.4$\to$\loss{79.9$\pm$2.7} \\
ccrcc & 76.7$\to$\gain{91.1$\pm$0.6} & 90.8$\to$90.5$\pm$0.5 & 93.6$\to$93.0$\pm$1.0 & 90.8$\to$90.9$\pm$0.3 & 89.5$\to$\gain{92.7$\pm$0.8} & 91.7$\to$\gain{92.2$\pm$0.3} & 88.7$\to$\gain{92.5$\pm$1.3} \\
crc & 92.0$\to$\gain{94.2$\pm$0.5} & 95.5$\to$\gain{96.2$\pm$0.2} & 92.5$\to$\gain{94.9$\pm$0.8} & 93.5$\to$\gain{94.7$\pm$0.2} & 94.1$\to$\gain{96.0$\pm$0.4} & 93.7$\to$\gain{96.0$\pm$0.2} & 94.8$\to$94.5$\pm$1.8 \\
esca & 77.3$\to$\gain{80.4$\pm$2.4} & 86.2$\to$87.1$\pm$1.6 & 88.9$\to$88.4$\pm$0.8 & 83.5$\to$\gain{85.8$\pm$0.4} & 86.4$\to$86.2$\pm$0.8 & 84.3$\to$84.8$\pm$1.2 & 80.1$\to$\gain{85.3$\pm$1.9} \\
mhist & 79.1$\to$\loss{78.7$\pm$0.2} & 80.3$\to$78.7$\pm$2.2 & 83.4$\to$84.1$\pm$3.5 & 83.3$\to$\loss{82.0$\pm$0.9} & 78.3$\to$\gain{80.8$\pm$1.7} & 82.4$\to$82.4$\pm$0.8 & 81.5$\to$81.9$\pm$2.3 \\
pcam & 90.8$\to$\gain{91.9$\pm$0.1} & 93.5$\to$\gain{94.1$\pm$0.2} & 92.8$\to$93.6$\pm$0.8 & 93.0$\to$93.5$\pm$0.6 & 95.0$\to$95.1$\pm$0.5 & 93.6$\to$94.0$\pm$0.8 & 92.6$\to$\gain{93.7$\pm$0.4} \\
spider\_br & 86.0$\to$\loss{84.6$\pm$0.5} & 86.1$\to$85.8$\pm$0.5 & 87.2$\to$85.8$\pm$1.5 & 87.2$\to$\loss{85.9$\pm$0.4} & 86.7$\to$\loss{86.3$\pm$0.3} & 86.2$\to$\loss{83.8$\pm$1.4} & 86.1$\to$\loss{85.1$\pm$0.8} \\
spider\_col & 89.7$\to$89.3$\pm$0.8 & 89.6$\to$89.7$\pm$0.2 & 90.8$\to$\loss{90.5$\pm$0.0} & 89.9$\to$90.0$\pm$0.1 & 90.5$\to$90.0$\pm$0.5 & 90.2$\to$90.1$\pm$0.1 & 89.1$\to$\loss{89.0$\pm$0.0} \\
spider\_sk & 87.2$\to$88.3$\pm$1.1 & 91.0$\to$\loss{90.1$\pm$0.2} & 92.0$\to$91.5$\pm$1.1 & 91.9$\to$\loss{91.3$\pm$0.1} & 92.5$\to$\loss{91.8$\pm$0.6} & 91.3$\to$\loss{90.3$\pm$0.4} & 90.1$\to$88.7$\pm$1.7 \\
spider\_th & 94.6$\to$93.8$\pm$1.4 & 94.4$\to$94.5$\pm$0.2 & 93.9$\to$93.2$\pm$1.0 & 94.4$\to$\loss{93.3$\pm$0.2} & 95.1$\to$94.8$\pm$0.6 & 94.7$\to$94.7$\pm$0.5 & 94.2$\to$93.6$\pm$0.9 \\
crc\_msi & 61.1$\to$\gain{62.6$\pm$0.6} & 65.6$\to$\gain{67.3$\pm$1.6} & 69.6$\to$\loss{68.7$\pm$0.3} & 67.3$\to$\loss{66.8$\pm$0.3} & 66.9$\to$68.1$\pm$1.3 & 65.5$\to$63.1$\pm$4.7 & 63.4$\to$\gain{67.1$\pm$1.6} \\
tils & 91.0$\to$90.9$\pm$0.2 & 91.0$\to$91.0$\pm$0.3 & 90.9$\to$90.4$\pm$1.3 & 91.0$\to$90.9$\pm$0.6 & 90.4$\to$90.7$\pm$0.3 & 90.7$\to$\gain{91.1$\pm$0.2} & 90.6$\to$\gain{91.3$\pm$0.0} \\
tcga\_unif & 77.7$\to$\loss{71.1$\pm$0.5} & 85.3$\to$\loss{83.1$\pm$2.1} & 78.4$\to$\loss{76.6$\pm$1.1} & 79.3$\to$\loss{77.6$\pm$0.6} & 78.8$\to$77.9$\pm$1.1 & 74.8$\to$73.9$\pm$1.1 & 80.3$\to$\loss{75.6$\pm$1.2} \\
wilds & 95.8$\to$\gain{97.0$\pm$1.0} & 98.3$\to$98.5$\pm$0.6 & 89.8$\to$\gain{98.6$\pm$0.1} & 98.7$\to$\loss{98.0$\pm$0.3} & 98.9$\to$98.9$\pm$0.1 & 97.6$\to$\gain{98.4$\pm$0.6} & 98.2$\to$\loss{91.6$\pm$1.4} \\
\midrule
\multicolumn{8}{l}{\emph{16-shot macro-F1\,$\uparrow$}} \\
\addlinespace[1pt]
bach & 57.3$\to$57.8$\pm$2.0 & 82.5$\to$79.6$\pm$4.2 & 80.7$\to$\loss{76.3$\pm$1.5} & 55.2$\to$\gain{63.6$\pm$1.6} & 84.7$\to$80.8$\pm$3.9 & 49.6$\to$\gain{60.9$\pm$3.7} & 76.0$\to$\loss{73.1$\pm$2.7} \\
bracs & 42.8$\to$\gain{46.6$\pm$0.4} & 49.6$\to$\gain{53.1$\pm$0.1} & 51.3$\to$\gain{55.1$\pm$0.1} & 43.1$\to$\gain{52.2$\pm$0.3} & 54.7$\to$54.8$\pm$2.3 & 40.1$\to$\gain{44.8$\pm$0.8} & 50.4$\to$\gain{54.8$\pm$0.6} \\
breakhis & 60.9$\to$\loss{58.0$\pm$1.3} & 38.5$\to$\gain{65.7$\pm$1.9} & 59.4$\to$\gain{66.3$\pm$3.0} & 70.6$\to$\loss{60.2$\pm$1.9} & 69.9$\to$\loss{67.8$\pm$1.5} & 50.5$\to$\gain{57.6$\pm$5.1} & 64.7$\to$\gain{68.8$\pm$2.3} \\
ccrcc & 90.1$\to$89.1$\pm$1.3 & 77.5$\to$\gain{85.4$\pm$2.3} & 79.9$\to$\gain{87.2$\pm$1.9} & 83.9$\to$\gain{85.6$\pm$1.0} & 90.6$\to$90.2$\pm$1.5 & 88.3$\to$\loss{86.5$\pm$0.5} & 63.6$\to$\gain{87.2$\pm$1.5} \\
crc & 89.6$\to$\gain{95.0$\pm$0.1} & 94.7$\to$\gain{95.9$\pm$0.1} & 91.3$\to$\gain{95.7$\pm$0.1} & 91.2$\to$\gain{94.6$\pm$0.1} & 95.5$\to$95.5$\pm$0.1 & 90.3$\to$\gain{94.3$\pm$0.5} & 94.7$\to$95.6$\pm$0.9 \\
esca & 67.0$\to$\loss{64.6$\pm$0.7} & 75.0$\to$\loss{73.8$\pm$0.3} & 70.0$\to$\gain{76.0$\pm$1.6} & 74.2$\to$\loss{73.3$\pm$0.5} & 76.7$\to$\loss{73.9$\pm$1.6} & 73.7$\to$\loss{67.6$\pm$1.8} & 75.9$\to$73.7$\pm$2.9 \\
mhist & 55.7$\to$\gain{62.1$\pm$4.2} & 62.4$\to$\gain{64.8$\pm$2.0} & 68.5$\to$\gain{70.0$\pm$1.4} & 65.5$\to$\gain{69.6$\pm$0.9} & 68.1$\to$68.7$\pm$1.1 & 60.0$\to$\gain{65.6$\pm$2.9} & 58.8$\to$\gain{68.5$\pm$1.8} \\
pcam & 83.5$\to$\loss{82.9$\pm$0.2} & 82.8$\to$\gain{86.8$\pm$1.6} & 80.7$\to$\gain{89.4$\pm$0.8} & 85.9$\to$\gain{88.1$\pm$0.8} & 92.9$\to$91.8$\pm$2.9 & 88.2$\to$\loss{87.1$\pm$0.2} & 85.3$\to$85.4$\pm$2.2 \\
spider\_br & 65.9$\to$67.1$\pm$1.4 & 61.4$\to$\gain{70.8$\pm$0.1} & 70.3$\to$\gain{74.1$\pm$1.2} & 76.4$\to$\loss{75.9$\pm$0.4} & 75.0$\to$\loss{73.7$\pm$0.6} & 75.6$\to$\loss{71.9$\pm$1.3} & 67.6$\to$68.4$\pm$1.3 \\
spider\_col & 82.3$\to$82.6$\pm$0.7 & 75.6$\to$\gain{83.7$\pm$0.6} & 76.7$\to$\gain{86.4$\pm$0.3} & 86.9$\to$\loss{86.2$\pm$0.1} & 86.0$\to$85.4$\pm$0.8 & 84.0$\to$\gain{84.6$\pm$0.2} & 82.0$\to$82.8$\pm$1.7 \\
spider\_sk & 75.4$\to$\gain{77.6$\pm$0.7} & 73.0$\to$\gain{78.0$\pm$1.1} & 79.2$\to$\gain{83.4$\pm$0.8} & 83.5$\to$83.7$\pm$0.9 & 85.0$\to$\loss{82.3$\pm$2.0} & 79.8$\to$\gain{81.0$\pm$0.6} & 76.0$\to$75.6$\pm$1.1 \\
spider\_th & 83.6$\to$\gain{88.5$\pm$0.8} & 86.4$\to$\gain{90.2$\pm$0.5} & 85.8$\to$\gain{90.7$\pm$0.4} & 91.5$\to$91.5$\pm$0.1 & 89.9$\to$\gain{91.8$\pm$0.2} & 87.8$\to$\gain{89.3$\pm$1.3} & 89.4$\to$88.9$\pm$1.7 \\
crc\_msi & 56.2$\to$\gain{58.6$\pm$0.2} & 55.1$\to$\gain{57.4$\pm$1.4} & 56.4$\to$\gain{62.2$\pm$0.6} & 61.4$\to$\gain{62.9$\pm$0.5} & 60.1$\to$61.2$\pm$1.1 & 57.0$\to$\gain{59.6$\pm$0.7} & 57.7$\to$\gain{60.1$\pm$0.7} \\
tils & 85.6$\to$\gain{86.6$\pm$0.2} & 76.2$\to$\gain{85.7$\pm$3.0} & 81.6$\to$\gain{84.2$\pm$1.3} & 88.5$\to$\gain{88.9$\pm$0.2} & 82.5$\to$\gain{83.5$\pm$0.3} & 76.9$\to$\gain{83.7$\pm$2.3} & 85.4$\to$\gain{90.4$\pm$0.2} \\
tcga\_unif & 60.2$\to$\loss{50.8$\pm$0.2} & 63.4$\to$\gain{66.2$\pm$2.6} & 57.4$\to$56.8$\pm$2.2 & 66.9$\to$\loss{60.2$\pm$0.9} & 69.2$\to$\loss{62.7$\pm$4.8} & 50.7$\to$\loss{48.1$\pm$1.4} & 63.5$\to$\loss{54.9$\pm$3.3} \\
wilds & 92.2$\to$92.2$\pm$1.1 & 89.9$\to$\gain{95.0$\pm$1.0} & 94.2$\to$\gain{95.2$\pm$0.9} & 93.6$\to$\gain{96.3$\pm$0.5} & 95.9$\to$\gain{96.4$\pm$0.1} & 96.8$\to$\loss{96.1$\pm$0.2} & 93.6$\to$93.7$\pm$2.0 \\
\bottomrule
\end{tabular}

%% file: tables/thunder_per_ds_other.tex
\begin{tabular}{lccccccc}
\toprule
Dataset & Phikon-v2 & Midnight-12k & Virchow2 & H-optimus-0 & UNI2-h & Virchow & OpenMidnight 1.1 \\
\midrule
\multicolumn{8}{l}{\emph{Segmentation F1\,$\uparrow$}} \\
\addlinespace[1pt]
ocelot & 79.3$\to$78.9$\pm$1.7 & 77.0$\to$79.2$\pm$5.0 & 79.6$\to$80.3$\pm$1.4 & 75.4$\to$77.3$\pm$2.0 & 80.8$\to$80.9$\pm$0.8 & 80.2$\to$79.5$\pm$1.3 & 79.8$\to$\gain{80.3$\pm$0.4} \\
pannuke & 60.4$\to$60.7$\pm$0.6 & 61.8$\to$62.2$\pm$0.5 & 62.9$\to$\loss{62.1$\pm$0.3} & 55.4$\to$54.9$\pm$1.8 & 62.0$\to$62.2$\pm$0.2 & 63.0$\to$62.5$\pm$0.7 & 62.5$\to$62.6$\pm$0.5 \\
segpath\_epi & 69.1$\to$\loss{68.6$\pm$0.1} & 70.5$\to$70.6$\pm$1.3 & 70.6$\to$70.4$\pm$0.4 & 69.3$\to$\loss{68.7$\pm$0.4} & 71.0$\to$71.1$\pm$0.5 & 70.8$\to$70.6$\pm$0.9 & 70.5$\to$70.4$\pm$0.8 \\
segpath\_lym & 60.4$\to$\gain{61.0$\pm$0.1} & 62.7$\to$62.7$\pm$0.2 & 62.7$\to$\gain{63.1$\pm$0.0} & 58.2$\to$\gain{58.9$\pm$0.2} & 62.9$\to$62.4$\pm$0.6 & 62.6$\to$62.8$\pm$0.9 & 63.8$\to$62.9$\pm$1.1 \\
\midrule
\multicolumn{8}{l}{\emph{Linear-probe ECE\,$\downarrow$}} \\
\addlinespace[1pt]
bach & 5.7$\to$8.4$\pm$4.1 & 5.6$\to$\loss{11.8$\pm$2.2} & 9.7$\to$\gain{7.2$\pm$1.7} & 13.3$\to$11.2$\pm$2.9 & 8.4$\to$\loss{23.4$\pm$3.8} & 20.3$\to$\gain{7.5$\pm$0.6} & 18.8$\to$\gain{8.4$\pm$6.3} \\
bracs & 7.6$\to$12.7$\pm$7.3 & 12.1$\to$\loss{14.4$\pm$1.9} & 6.7$\to$\loss{20.2$\pm$1.4} & 14.5$\to$12.9$\pm$2.8 & 6.2$\to$9.6$\pm$3.9 & 11.5$\to$\loss{15.8$\pm$0.8} & 15.2$\to$\gain{11.4$\pm$1.6} \\
breakhis & 17.2$\to$14.0$\pm$3.5 & 2.1$\to$5.8$\pm$4.7 & 5.7$\to$\loss{12.8$\pm$1.6} & 2.9$\to$4.4$\pm$3.3 & 6.9$\to$8.1$\pm$3.5 & 12.8$\to$12.2$\pm$4.1 & 7.7$\to$8.6$\pm$3.8 \\
ccrcc & 9.3$\to$\gain{6.4$\pm$0.3} & 3.5$\to$1.9$\pm$2.6 & 4.3$\to$\gain{3.5$\pm$0.6} & 6.2$\to$\gain{4.0$\pm$0.1} & 8.1$\to$\gain{3.7$\pm$1.9} & 2.0$\to$\loss{2.7$\pm$0.3} & 22.2$\to$\gain{7.2$\pm$5.4} \\
crc & 4.8$\to$\gain{3.4$\pm$0.2} & 2.7$\to$\gain{2.3$\pm$0.1} & 4.1$\to$\gain{3.0$\pm$0.3} & 4.1$\to$\gain{3.2$\pm$0.1} & 3.8$\to$\gain{2.4$\pm$0.3} & 3.8$\to$\gain{2.5$\pm$0.5} & 3.3$\to$3.1$\pm$0.7 \\
esca & 0.8$\to$1.3$\pm$0.8 & 0.9$\to$0.9$\pm$0.8 & 0.5$\to$0.7$\pm$0.8 & 2.9$\to$\gain{0.6$\pm$0.0} & 1.2$\to$0.8$\pm$0.5 & 0.5$\to$\loss{0.8$\pm$0.1} & 3.4$\to$\gain{0.5$\pm$0.1} \\
mhist & 2.5$\to$2.3$\pm$1.7 & 2.1$\to$2.9$\pm$2.8 & 1.2$\to$3.7$\pm$5.1 & 1.6$\to$3.6$\pm$2.8 & 7.2$\to$6.1$\pm$5.8 & 3.4$\to$2.7$\pm$0.9 & 4.1$\to$5.5$\pm$3.0 \\
pcam & 0.6$\to$\loss{1.6$\pm$0.0} & 0.7$\to$\gain{0.3$\pm$0.2} & 4.3$\to$\gain{3.1$\pm$0.7} & 2.9$\to$2.6$\pm$1.1 & 1.2$\to$\gain{0.4$\pm$0.5} & 1.5$\to$1.8$\pm$1.2 & 2.1$\to$\gain{0.8$\pm$0.3} \\
spider\_br & 2.2$\to$2.0$\pm$4.7 & 1.7$\to$3.3$\pm$1.8 & 3.1$\to$4.0$\pm$2.5 & 3.4$\to$3.9$\pm$3.6 & 5.8$\to$\gain{3.2$\pm$1.6} & 4.8$\to$5.0$\pm$1.1 & 3.4$\to$\loss{4.0$\pm$0.4} \\
spider\_col & 2.4$\to$\gain{0.3$\pm$0.3} & 1.1$\to$1.1$\pm$1.0 & 0.3$\to$\loss{1.0$\pm$0.3} & 1.1$\to$0.8$\pm$0.4 & 0.3$\to$\loss{0.5$\pm$0.1} & 0.5$\to$\gain{0.3$\pm$0.1} & 2.3$\to$\gain{0.5$\pm$0.2} \\
spider\_sk & 2.5$\to$2.1$\pm$2.2 & 3.2$\to$3.2$\pm$1.6 & 0.9$\to$1.1$\pm$0.7 & 0.3$\to$0.4$\pm$0.2 & 2.0$\to$\gain{0.9$\pm$0.3} & 1.0$\to$\loss{2.6$\pm$0.4} & 2.1$\to$\loss{3.2$\pm$0.6} \\
spider\_th & 0.5$\to$1.2$\pm$1.7 & 1.5$\to$\gain{0.3$\pm$0.3} & 2.7$\to$2.8$\pm$0.7 & 2.0$\to$\gain{1.2$\pm$0.5} & 0.4$\to$0.8$\pm$0.7 & 0.5$\to$0.6$\pm$0.5 & 0.6$\to$\loss{1.7$\pm$0.2} \\
crc\_msi & 0.9$\to$2.6$\pm$3.9 & 2.5$\to$\loss{5.1$\pm$1.7} & 5.5$\to$4.6$\pm$1.5 & 3.1$\to$\loss{4.8$\pm$0.1} & 5.0$\to$\loss{8.9$\pm$3.1} & 2.6$\to$5.4$\pm$5.3 & 4.7$\to$4.8$\pm$5.2 \\
tils & 0.8$\to$0.9$\pm$0.3 & 0.8$\to$\gain{0.3$\pm$0.4} & 0.3$\to$1.3$\pm$1.4 & 0.5$\to$0.5$\pm$0.9 & 1.1$\to$\gain{0.4$\pm$0.2} & 0.8$\to$0.5$\pm$0.6 & 0.4$\to$0.3$\pm$0.3 \\
tcga\_unif & 0.8$\to$\loss{2.7$\pm$0.3} & 5.6$\to$6.5$\pm$1.8 & 7.2$\to$\gain{2.0$\pm$0.4} & 3.1$\to$1.5$\pm$2.4 & 3.6$\to$2.5$\pm$2.7 & 5.8$\to$\gain{3.1$\pm$0.2} & 1.8$\to$3.6$\pm$3.1 \\
wilds & 1.6$\to$\gain{0.3$\pm$0.3} & 0.2$\to$0.6$\pm$0.9 & 7.1$\to$\gain{0.2$\pm$0.2} & 1.9$\to$\gain{0.2$\pm$0.1} & 1.0$\to$\gain{0.5$\pm$0.3} & 0.6$\to$0.6$\pm$0.3 & 0.2$\to$\loss{6.5$\pm$1.2} \\
\midrule
\multicolumn{8}{l}{\emph{Adversarial macro-F1 drop\,$\downarrow$ (fp32-attack cells)}} \\
\addlinespace[1pt]
bach & 43.7$\to$\gain{33.3$\pm$1.6} & 39.2$\to$\gain{18.1$\pm$4.3} & 20.5$\to$\gain{11.9$\pm$1.2} & 40.8$\to$\gain{29.3$\pm$4.8} & 27.7$\to$20.4$\pm$7.5 & 38.8$\to$\gain{29.7$\pm$2.4} & 32.1$\to$\gain{21.9$\pm$3.9} \\
bracs & 41.3$\to$\gain{32.8$\pm$6.0} & 47.2$\to$\gain{39.4$\pm$3.6} & 31.7$\to$34.8$\pm$5.6 & 43.7$\to$\gain{36.2$\pm$0.9} & 40.3$\to$\gain{32.6$\pm$2.6} & 40.1$\to$\gain{34.2$\pm$4.6} & 37.1$\to$\gain{33.3$\pm$1.4} \\
breakhis & 28.2$\to$29.7$\pm$4.7 & 38.3$\to$41.4$\pm$11.7 & 25.8$\to$\gain{22.6$\pm$2.3} & 28.0$\to$\loss{32.6$\pm$0.4} & 29.2$\to$\gain{21.8$\pm$6.9} & 42.8$\to$\gain{36.8$\pm$4.0} & 52.4$\to$\gain{26.2$\pm$3.4} \\
ccrcc & 37.6$\to$\gain{20.5$\pm$1.7} & 31.2$\to$\gain{19.6$\pm$5.6} & 19.3$\to$\gain{12.9$\pm$0.8} & 33.3$\to$\gain{20.5$\pm$1.0} & 23.9$\to$\gain{15.4$\pm$0.7} & 28.7$\to$\gain{18.4$\pm$0.5} & 43.1$\to$\gain{13.6$\pm$0.8} \\
crc & 44.7$\to$\gain{38.9$\pm$1.9} & 29.8$\to$\gain{20.0$\pm$5.2} & 23.2$\to$\gain{19.9$\pm$2.9} & 30.7$\to$\gain{25.0$\pm$1.3} & 30.7$\to$\gain{19.2$\pm$1.9} & 28.0$\to$\gain{21.7$\pm$4.1} & 28.4$\to$\gain{19.4$\pm$2.8} \\
esca & 29.3$\to$\gain{26.2$\pm$2.6} & 28.8$\to$\gain{19.0$\pm$3.9} & 17.2$\to$17.1$\pm$4.3 & 28.6$\to$\gain{18.8$\pm$0.5} & 18.4$\to$\gain{16.5$\pm$1.2} & 20.9$\to$20.2$\pm$1.5 & 20.0$\to$18.5$\pm$3.7 \\
mhist & 64.8$\to$\gain{58.5$\pm$0.3} & 61.4$\to$55.8$\pm$7.7 & 42.1$\to$43.5$\pm$8.8 & 62.1$\to$59.3$\pm$3.8 & 61.3$\to$57.3$\pm$11.0 & 63.9$\to$\gain{52.0$\pm$4.2} & 58.2$\to$58.1$\pm$2.5 \\
pcam & 67.0$\to$\gain{54.1$\pm$8.3} & 57.4$\to$\gain{41.4$\pm$15.4} & 58.2$\to$\gain{31.2$\pm$9.0} & 100.0$\to$\gain{86.0$\pm$10.1} & 42.5$\to$\gain{29.0$\pm$2.6} & 65.0$\to$\gain{39.9$\pm$3.3} & 52.5$\to$\gain{31.8$\pm$4.9} \\
spider\_br & 50.3$\to$41.3$\pm$13.6 & 49.1$\to$\gain{36.1$\pm$4.6} & 34.7$\to$31.9$\pm$7.8 & 58.0$\to$\gain{49.4$\pm$8.4} & 38.1$\to$\gain{28.4$\pm$3.6} & 42.3$\to$\gain{32.2$\pm$1.6} & 39.7$\to$\gain{31.6$\pm$3.1} \\
spider\_col & 37.1$\to$\gain{29.2$\pm$1.9} & 36.6$\to$\gain{24.6$\pm$3.4} & 19.6$\to$\gain{16.3$\pm$1.7} & 33.6$\to$\gain{25.0$\pm$1.0} & 21.4$\to$\gain{17.3$\pm$0.4} & 27.3$\to$\gain{20.0$\pm$2.3} & 34.2$\to$\gain{20.4$\pm$2.5} \\
spider\_sk & 36.4$\to$\gain{29.1$\pm$3.1} & 38.9$\to$\gain{28.2$\pm$0.3} & 21.6$\to$\gain{18.9$\pm$0.5} & 38.2$\to$\gain{30.9$\pm$0.6} & 21.2$\to$19.9$\pm$2.0 & 26.4$\to$\gain{23.5$\pm$0.4} & 31.1$\to$\gain{23.0$\pm$2.1} \\
spider\_th & 30.7$\to$25.5$\pm$5.9 & 32.3$\to$\gain{20.7$\pm$5.2} & 18.7$\to$18.0$\pm$2.8 & 42.4$\to$\gain{31.9$\pm$3.5} & 14.5$\to$14.8$\pm$6.3 & 24.6$\to$23.3$\pm$2.4 & 24.0$\to$\gain{19.6$\pm$2.8} \\
crc\_msi & 38.8$\to$\gain{30.9$\pm$2.5} & 26.3$\to$\gain{17.1$\pm$0.4} & 29.8$\to$\gain{24.8$\pm$1.2} & 40.5$\to$\gain{30.5$\pm$0.4} & 31.1$\to$\gain{22.4$\pm$4.7} & 38.2$\to$\gain{31.4$\pm$1.4} & 25.3$\to$21.0$\pm$4.3 \\
tils & 35.5$\to$35.3$\pm$2.3 & 19.5$\to$24.2$\pm$6.8 & 43.0$\to$\gain{39.4$\pm$1.6} & 43.3$\to$\loss{46.0$\pm$1.0} & 40.3$\to$\gain{32.8$\pm$4.1} & 34.9$\to$\gain{31.7$\pm$0.9} & 24.6$\to$\gain{22.6$\pm$0.9} \\
tcga\_unif & 52.5$\to$\gain{43.7$\pm$0.8} & 36.7$\to$37.3$\pm$2.2 & 43.5$\to$\gain{36.8$\pm$2.1} & 51.7$\to$\gain{45.1$\pm$1.8} & 40.9$\to$\gain{38.6$\pm$1.3} & 52.9$\to$\gain{45.9$\pm$1.5} & 41.3$\to$38.2$\pm$3.1 \\
wilds & 60.2$\to$\gain{29.0$\pm$2.8} & 22.7$\to$23.2$\pm$23.9 & 51.5$\to$\gain{25.8$\pm$3.6} & 31.0$\to$\loss{36.3$\pm$1.6} & 24.1$\to$\gain{12.4$\pm$1.0} & 39.2$\to$\gain{24.4$\pm$3.3} & 12.9$\to$\loss{21.3$\pm$2.9} \\
\bottomrule
\end{tabular}

%% file: tables/perseed_thunder.tex
\begin{tabular}{lcccccccccccccccccccc}
\toprule
 & \multicolumn{4}{c}{$k$-NN F1} & \multicolumn{4}{c}{LP F1} & \multicolumn{4}{c}{16-shot F1} & \multicolumn{4}{c}{Seg.\ F1} & \multicolumn{4}{c}{LP ECE} \\
\cmidrule(lr){2-5}\cmidrule(lr){6-9}\cmidrule(lr){10-13}\cmidrule(lr){14-17}\cmidrule(lr){18-21}
Backbone & Base & Seed a & Seed b & Seed c & Base & Seed a & Seed b & Seed c & Base & Seed a & Seed b & Seed c & Base & Seed a & Seed b & Seed c & Base & Seed a & Seed b & Seed c \\
\midrule
Phikon-v2 & 73.9 & 77.3 & 77.7 & 77.3 & 79.7 & 81.4 & 81.2 & 81.5 & 71.8 & 72.5 & 72.5 & 72.5 & 67.3 & 67.4 & 67.0 & 67.4 & 3.8 & 3.8 & 3.8 & 4.0 \\
Midnight-12k & 80.0 & 81.4 & 82.1 & 81.8 & 84.8 & 85.7 & 85.5 & 85.7 & 71.5 & 77.0 & 77.1 & 76.9 & 68.0 & 68.1 & 69.0 & 69.0 & 2.9 & 3.5 & 4.0 & 3.9 \\
Virchow2 & 82.9 & 83.0 & 83.0 & 83.2 & 84.7 & 85.2 & 85.6 & 85.6 & 74.0 & 78.2 & 77.7 & 78.3 & 69.0 & 69.2 & 68.8 & 69.0 & 4.0 & 4.7 & 4.3 & 4.4 \\
H-optimus-0 & 81.4 & 81.6 & 81.2 & 81.7 & 83.8 & 84.2 & 84.1 & 84.3 & 76.1 & 77.0 & 77.0 & 77.1 & 64.6 & 64.8 & 64.6 & 65.5 & 4.0 & 3.5 & 3.4 & 3.6 \\
UNI2-h & 83.3 & 83.3 & 82.5 & 83.1 & 85.7 & 86.1 & 85.5 & 85.8 & 79.8 & 79.2 & 78.2 & 78.9 & 69.2 & 69.0 & 69.0 & 69.4 & 3.9 & 4.6 & 4.4 & 4.5 \\
Virchow & 77.4 & 79.2 & 79.0 & 78.6 & 82.8 & 83.6 & 83.4 & 83.2 & 71.8 & 73.6 & 73.9 & 73.5 & 69.2 & 69.0 & 68.8 & 68.8 & 4.5 & 3.9 & 3.8 & 4.3 \\
OpenMidnight 1.1 & 80.8 & 80.5 & 81.4 & 81.5 & 85.2 & 84.7 & 84.9 & 84.7 & 74.0 & 75.9 & 76.7 & 76.5 & 69.2 & 69.1 & 68.8 & 69.2 & 5.8 & 4.2 & 4.6 & 4.3 \\
\bottomrule
\end{tabular}

%% file: tables/ablations.tex
\begin{tabular}{lcccccc}
\toprule
\multicolumn{7}{l}{\emph{(a) Pseudo-tissue reweighting: asymmetric $(+3,-\infty)$ minus none $(0,0)$, seed 0, at the asymmetric arm's selected checkpoint}} \\
Backbone & $\Delta$RI & $\Delta$HEST & Tuned HEST $2$\,SD & & & \\
\cmidrule(lr){1-4}
Phikon-v2    & $+0.0091$          & $+0.0044$ & $0.0006$ & & & \\
Midnight-12k & $-0.0002^\dagger$  & $+0.0016$ & $0.0014$ & & & \\
Virchow2     & $+0.0030^\dagger$  & $-0.0005$ & $0.0034$ & & & \\
H-optimus-0  & $+0.0003^\dagger$  & $+0.0020$ & $0.0018$ & & & \\
UNI2-h       & $+0.0004^\dagger$  & $-0.0007$ & $0.0058$ & & & \\
\midrule
\multicolumn{7}{l}{\emph{(b) Batch composition, Phikon-v2, one seed, 750-step sweep at a fixed 3,600-image budget ($CT$ constant)}} \\
$C$ & RI @750 & & Change at fixed $C$ & $\Delta$RI @750 & @1000 & @1250 \\
\cmidrule(lr){1-2}\cmidrule(lr){4-7}
24 & 0.802          & & $C{=}24$, $T$ 100$\to$150 & $-0.0002$ & $-0.0033$ & $-0.0063$ \\
12 & 0.809          & & $C{=}12$, $T$ 200$\to$300 & $+0.0060$ & $+0.0036$ & $+0.0026$ \\
6  & 0.814          & & & & & \\
2  & \textbf{0.824} & & & & & \\
\midrule
\multicolumn{7}{l}{\emph{(c) Adapter rank at fixed $\alpha/r=2$, one seed: mean RI over checkpoints at the stated steps}} \\
Backbone & Steps & $r{=}8$ & $r{=}16$ & $r{=}32$ & $r{=}64$ & $r{=}128$ \\
\cmidrule(lr){1-7}
Phikon-v2 & $\ge1500$ & 0.7998 & 0.8019 & \textbf{0.8029} & 0.7969 & 0.8009 \\
Virchow2  & $\ge750$  & --     & --     & \textbf{0.9006} & 0.8983 & 0.8735 \\
\bottomrule
\end{tabular}